%% file: main.tex
\documentclass[10pt,twocolumn,letterpaper]{article}

\usepackage[pagenumbers]{cvpr} 


\definecolor{cvprblue}{rgb}{0.21,0.49,0.74}
\usepackage[pagebackref,breaklinks,colorlinks,allcolors=cvprblue]{hyperref}
\usepackage{ulem}
\usepackage{adjustbox}
\usepackage{multirow}
\usepackage{threeparttable}
\usepackage{pifont}
\usepackage{xspace}
\usepackage{colortbl}
\usepackage{xcolor}
\usepackage{arydshln}
\usepackage{amsmath, amssymb, amsfonts}
\usepackage{mathtools}   
\usepackage{bm} 
\usepackage{placeins}
\def\paperID{11} 
\def\confName{CVPR}
\def\confYear{2026}

\title{Iris: \underline{I}nteg\underline{r}ating Language \underline{i}nto Diffu\underline{s}ion-based Monocular Depth Estimation}

\newcommand{\paper}{Iris\xspace}

\author{Ziyao Zeng\textsuperscript{*1} \quad
Jingcheng Ni\textsuperscript{*2} \quad
Daniel Wang\textsuperscript{1} \quad
Patrick Rim\textsuperscript{1} \quad \\
Younjoon Chung\textsuperscript{1} \quad
Fengyu Yang\textsuperscript{1} \quad
Byung-Woo Hong\textsuperscript{3} \quad
Alex Wong\textsuperscript{1} \quad
\vspace{3mm} \\
\textsuperscript{1}Yale University \quad \textsuperscript{2}Brown University \quad Chung-Ang University\textsuperscript{3} \\ 
\tt\small \textsuperscript{1}\{ziyao.zeng,
daniel.wang.dhw33,
patrick.rim,
fengyu.yang\}@yale.edu \\
\tt\small \textsuperscript{1}\{younjoon.chung,
alex.wong\}@yale.edu 
\tt\small  \textsuperscript{2}jingcheng\_ni@brown.edu
\tt\small  \textsuperscript{3}hong@cau.ac.kr}

\begin{document}
\maketitle



\begingroup
\renewcommand\thefootnote{}
\footnotetext{* Equal contribution.}
\endgroup

\begin{abstract}

Traditional monocular depth estimation suffers from inherent ambiguity and visual nuisances. We demonstrate that language can enhance monocular depth estimation by providing an additional condition (rather than images alone) aligned with plausible 3D scenes, thereby reducing the solution space for depth estimation. This conditional distribution is learned during the text-to-image pre-training of diffusion models. To generate images under various viewpoints and layouts that precisely reflect textual descriptions, the model implicitly models object sizes, shapes, and scales, their spatial relationships, and the overall scene structure. In this paper, \paper, we investigate the benefits of our strategy to integrate text descriptions into training and inference of diffusion-based depth estimation models. We experiment with three different diffusion-based monocular depth estimators (Marigold, Lotus, and E2E-FT) and their variants. By training on HyperSim and Virtual KITTI, and evaluating on NYUv2, KITTI, ETH3D, ScanNet, and DIODE, we find that our strategy improves the overall monocular depth estimation accuracy, especially in small areas. It also improves the model's depth perception of specific regions described in the text. We find that by providing more details in the text, the depth prediction can be iteratively refined. Simultaneously, we find that language can act as a constraint to accelerate the convergence of both training and the inference diffusion trajectory. Code and generated text data will be released upon acceptance.

\end{abstract}

\begin{figure}[t]
  \centering
  \vspace{0.3cm}
\includegraphics[width=0.45\textwidth]{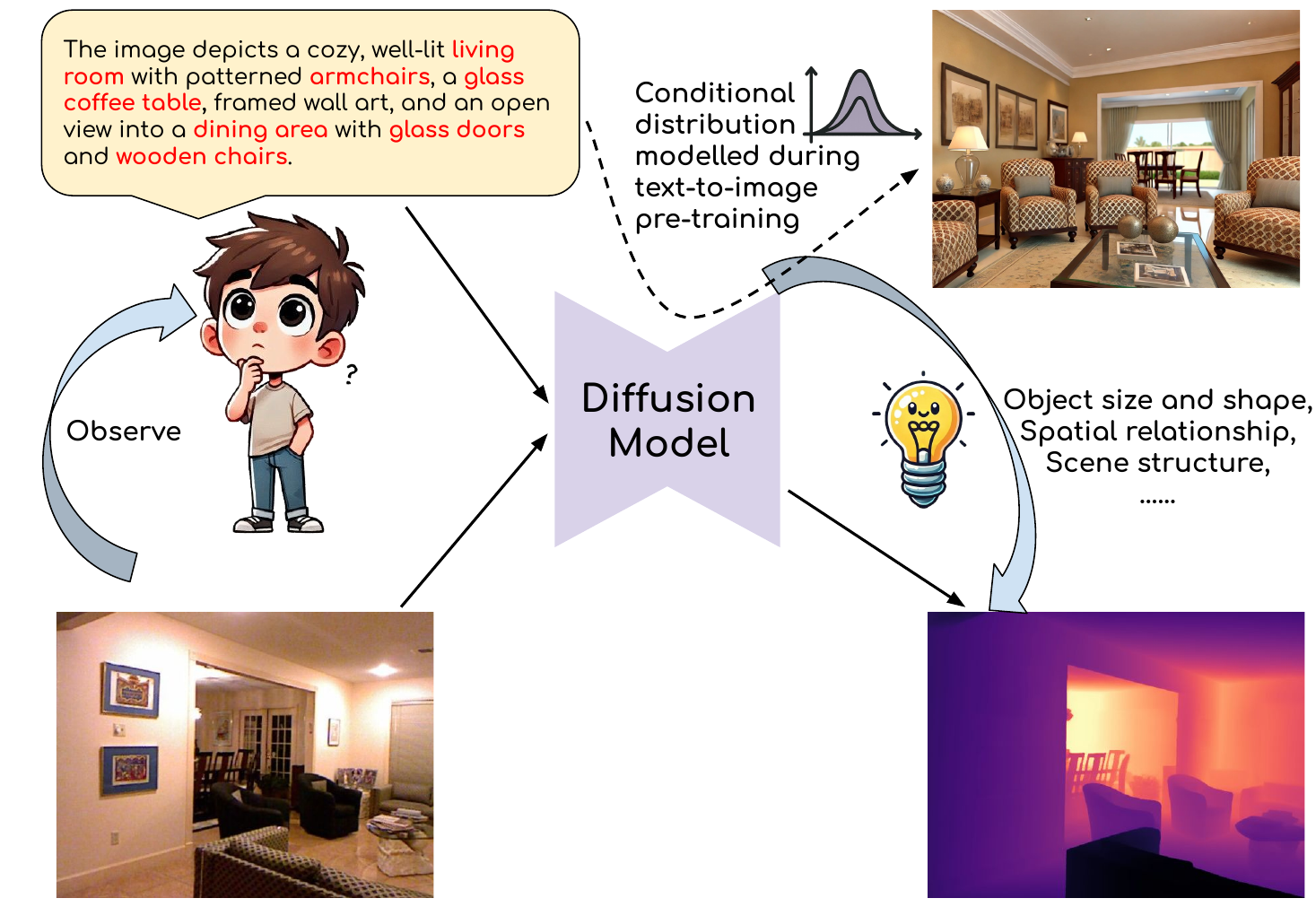}
    \caption{
        \textbf{Integrating language into diffusion models enhances monocular depth estimation} by providing an additional condition (rather than images alone) associated with plausible 3D scenes, thus reducing the solution space for depth estimation. This conditional distribution is initially learned during text-to-image generation pre-training of diffusion models, as to generate images under different viewpoints and layouts that accurately reflect the text, the model needs to implicitly model the size and shape of specified objects, their spatial relationship, and the structure of the scene. Then the conditional distribution is associated with plausible 3D scenes during fine-tuning with image-text-depth pairs.
    }
    \label{fig:teaser}
    \vspace{-0.5cm}
\end{figure}


\section{Introduction}
\label{sec:intro}
Monocular depth estimation requires the model to predict pixel-wise depth from a single image. Recent advancements in diffusion-based models have shown promise for generating high-quality depth maps~\cite{duan2023diffusiondepth,ji2023ddp,saxena2023monocular,saxena2024surprising,zhao2023unleashing, Marigold, fu2024geowizard, gui2024depthfm, fei2019geo, garcia2024fine, he2024lotus, xu2024diffusion} due to their ability to capture complex structures and fine details by progressively denoising a latent representation to produce accurate depth predictions. However, it remains a challenging task due to inherent ambiguities and various visual nuisances. Unlike stereo or multi-view depth estimation methods, monocular approaches lack visual features provided by multiple perspectives, leading to difficulty in accurately discerning the relative sizes,  distances, and spatial relationships of objects within a scene. That is, different real-world 3D scenes can project to the same 2D image, causing inherent ambiguities. For example, texture ambiguity. Surfaces that are uniformly textured or have repetitive patterns, like tiled floors, can create confusion for depth estimators. The models might misinterpret the repetition as uniform depth, even though the actual depth might vary (e.g., a tiled floor receding into the distance). Also, visibility ambiguity creates limitations for discerning small, partial or occluded objects purely from the visual signal. On the other hand, visual nuisances such as illumination,  motion blur, atmospheric conditions, and perspective distortion distort or obscure critical visual cues, leading vision algorithms to misinterpret scene structure, depth, and object boundaries, thus reducing their accuracy and reliability.

\begin{figure}[t!]
    \includegraphics[width=\linewidth]{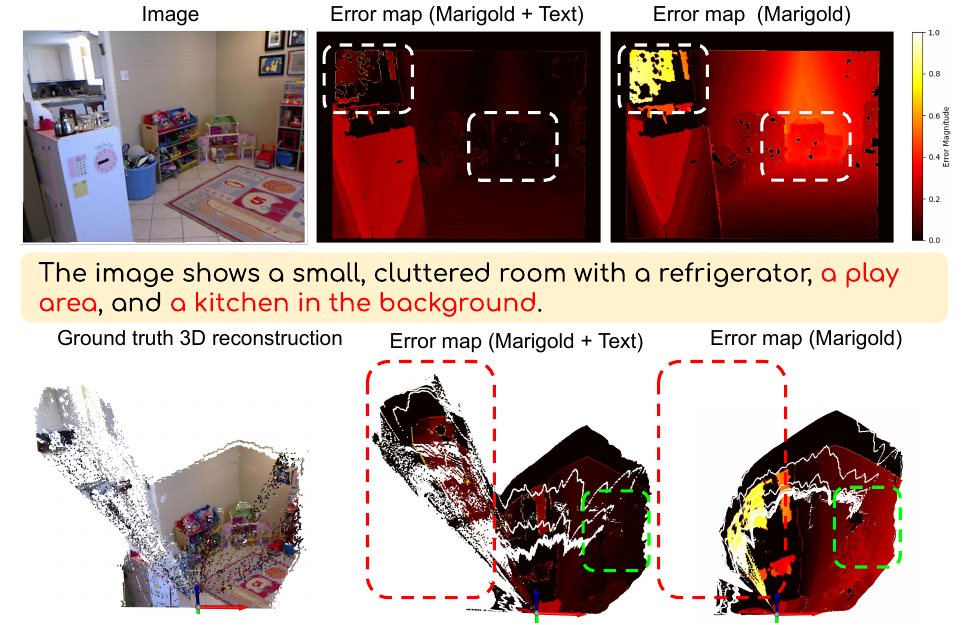}
    \caption{\textbf{Language improves the depth perception of specified insignificant (and potentially ambiguous) regions.}}
    \label{fig:3D_recon}
    \vspace{-5mm}
\end{figure}

Shown in Figure~\ref{fig:teaser}, an emerging solution lies in integrating language as a complementary modality to provide an additional condition to resolve ambiguity and visual nuisance. Language descriptions provided by humans offer a wealth of conditions about spatial relationships, scene types, object sizes, and depth hierarchies. 
In cases where an object is small, partially occluded, visually similar to the background, or its type is ambiguous from a single image, language can indicate the object's identity and semantics, enabling the model to infer its typical 3D size and shape and make more precise depth predictions for the observed portion.
Language inputs are also valuable in resolving texture ambiguities. Descriptive cues can specify the type and function of surfaces in a scene, aiding the model in inferring geometric properties. For instance, if a uniformly textured surface is described as a ``white ceiling,'' the model can interpret it as a continuous, flat plane at a consistent depth above the observer, extending into the distance. Similarly, for surfaces with repetitive patterns, such as a red and black interwoven carpet, language inputs like ``A red and black interwoven carpet extending across the room'' can help the model recognize that the repeating pattern belongs to a flat surface, preventing misinterpretation of edges between tiles as depth discontinuities.
As shown in Figure~\ref{fig:3D_recon}, better depth perception can be achieved for regions specified in the language description. Without language, the baseline model misinterprets the background kitchen as a flat surface, likely a wallpaper or a painting. With the additional condition ``a kitchen in the background,'' the model applies an additional condition about the structure of a kitchen, leading to more accurate depth prediction. In real-world applications of embodied agents, human provides text instruction to agents, which contains the description of the scene and can be used to enhance models' 3D perception at no additional cost. In the example of Figure~\ref{fig:3D_recon}, if an agent is instructed to grab something ``in the kitchen in the background'', the instruction can be integrated into the 3D perception model to avoid treating the background as a flat surface. Also, when generalizing to a new environment, language, having less covariate shift compared to images~\cite{zeng2024rsa, zeng2024wordepth}, allows humans to describe the new environment, aiding agents in better adapting and generalizing to it.


To this end, we investigate incorporating language descriptions of the scene as conditions to enhance the depth map prediction. During text-to-image pre-training, diffusion models learn to generate diverse images under various viewpoints and scene layouts that align with the provided language descriptions. The ability of these models to generate images that align with text suggests these models are implicitly modeling the spatial relationships, size, shape, and scales of specified objects, as well as the structure of the 3D scene, then associate such implicit modeling with text. When integrating text into the training and inference of diffusion-based monocular depth estimator, text is used as a condition to reduce the solution space of possible 3D scenes, thus improve the accuracy of depth prediction. In our implementation, during the denoising process, both the image and the text input are used by the model as conditions to predict the noise to be removed. Ultimately, Gaussian noise is progressively refined into a depth map that aligns with both the input image and the language description.

To support our hypothesis, we examine three different diffusion-based monocular depth estimator and their variants: Marigold~\cite{Marigold}, Lotus~\cite{he2024lotus}, and E2E-FT~\cite{garcia2024fine}. We conduct training on two synthetic datasets, HyperSim~\cite{roberts2021hypersim} and Virtual KITTI~\cite{VirtualKITTI}, and conduct zero-shot evaluation on five real-world datasets, NYUv2~\cite{NYU-Depth-V2}, KITTI~\cite{KITTI}, ETH3D~\cite{ETH3D}, ScanNet~\cite{ScanNet}, and DIODE~\cite{DIODE}. Given the difficulty of obtaining sufficient human-provided text descriptions for every training and inference image, we utilize a vision-language model, i.e. LLaVA v1.6~\cite{liu2023improved} and InternVL3-8B~\cite{zhu2025internvl3}, to generate descriptions for each image, simulating human annotation. By integrating text into both the training and inference stages of a diffusion-based depth estimator, we find that it enhances overall monocular depth estimation accuracy and improves the model’s depth perception in regions described by the text. Furthermore, providing more detailed textual descriptions enables iterative refinement of the predicted depth. At the same time, language serves as a constraint that accelerates the convergence of both training and the diffusion trajectory during inference. Interestingly, we also observe that in some cases, even when text is only provided in training or inference time, the accuracy of monocular depth estimation still shows a slight improvement. One possible explanation is that when the model is trained with text but evaluated without it, the training-time text descriptions help the model disambiguate challenging or visually ambiguous regions. By learning these associations, the model develops a stronger correspondence between such regions and their ground-truth depth, which in turn improves depth prediction even when text is absent at inference time. Conversely, when the model is trained without text but evaluated with it, the diffusion model can still partially utilize text features. Although no text–depth correspondence is learned during training, text prompts can activate conditions that diffusion models have learned from large-scale text-to-image pre-training—such as associations with appearance, structure, or texture. These implicit priors can provide weak but useful guidance, thereby improving depth estimation to some extent even without explicit text–depth supervision.

\noindent
\textbf{Our findings.} We summarize our findings of the influence of integrating language into the diffusion-based monocular depth estimator as below:

\begin{itemize}
    \item \textbf{Finding 1:} Language provides a condition about the existence, geometric properties, and spatial relationships of objects and scene structures, helping depth estimators to reduce the depth solution space to better perceive depth, especially in insignificant or ambiguous regions.
    \item \textbf{Finding 2:} Depth prediction can be iteratively refined with more details in the text description. This is particularly beneficial for regions that pose challenges to vision systems, such as those with small size, poor illumination, occlusion, or high visual similarity to the background.
    \item \textbf{Finding 3:} Language serves as a constraint that accelerates training convergence and provides a good initialization of the diffusion trajectory to speed up inference. This improves the efficiency of the diffusion model.
\end{itemize}




\begin{figure*}[t!]
  \centering
    \includegraphics[width=1.0\textwidth]{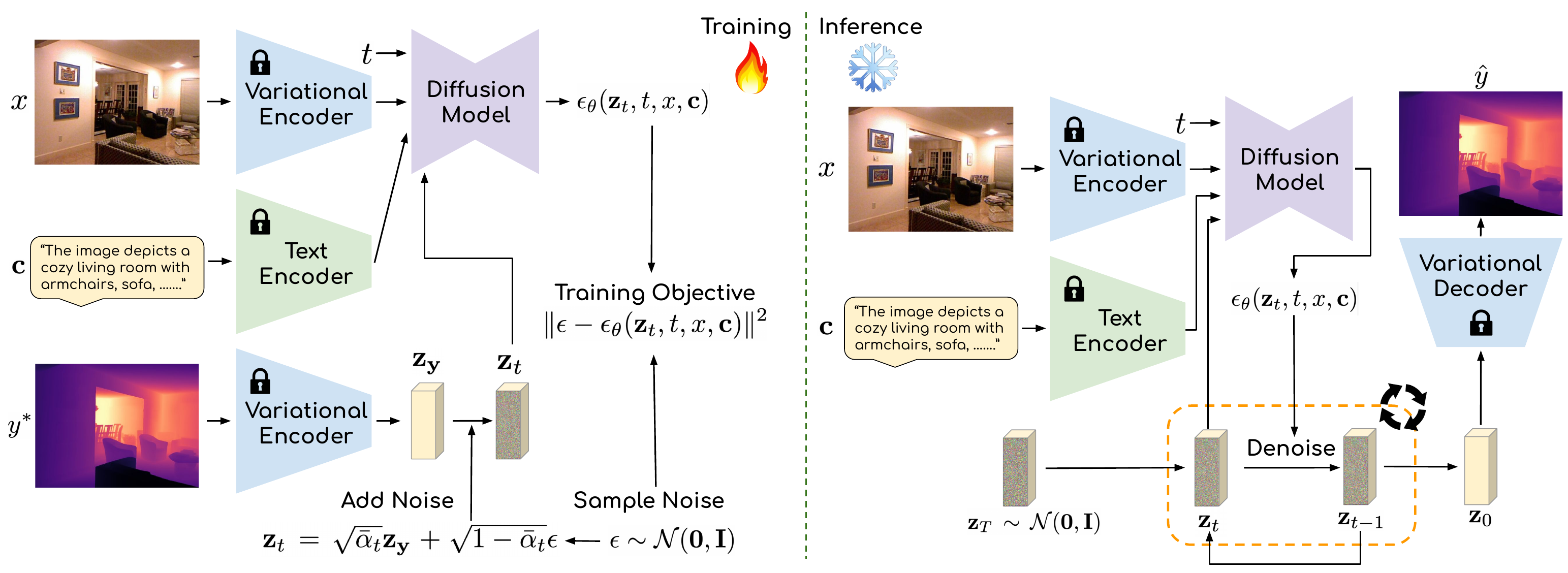}   \caption{\textbf{Pipeline to integrate text.} We train the diffusion model to predict the noise added into the noisy depth latent $\mathbf{z}_t$ at the time step $t$, based on $\mathbf{z}_t$, the input image $x$, and the corresponding language description $c$. During inference, the diffusion model predicts noise for $\mathbf{z}_t$ at each time step and gradually denoise it from $\mathbf{z}_T$ (pure Gaussian noise) into $\mathbf{z}_0$ (pure depth latent). Then $\mathbf{z}_0$ is decoded into the depth prediction using a frozen variational decoder.}
    \label{fig:pipeline}
    \vspace{-0.3cm}
\end{figure*}
\section{Related Work}
\label{sec:related_work}
\textbf{Monocular depth estimation.} 
Monocular depth estimation is a dense regression problem, requiring pixel-wise predictions of depth from a single image, capturing complex spatial relationships within the scene. Some depth models learn to infer pixel-wise depth in metric scale (i.e. meters)~\cite{bhat2021adabins,dorn,lao2024sub,Va-depthnet,wong2020targeted,yuan2022neural,DepthCLIP,zeng2024wordepth, chung2025eta, rim2025protodepth} by minimizing loss between depth predictions and ground-truth depth maps. Each model typically applies to only one data domain in which it is trained, with similar camera parameters and object scales.
To enable generalization across diverse domains, empowering depth estimation in the wild, other works have explored affine-invariant depth estimation~\cite{zhang2022hierarchical, DepthAnything, DPT, midas, Marigold, DepthAnythingV2,yin2021learning,eftekhar2021omnidata,yin2020diversedepth,li2018megadepth}, which predicts depth up to an unknown global shift and scale. This approach can accommodate varied scenes with different scales, while still preserving the geometric relationships between different objects or regions within the scene.
HND~\cite{zhang2022hierarchical} hierarchically normalizes the depth representations with spatial information and depth distributions. 
Depth Anything ~\cite{DepthAnything} learns from large-scale automatically annotated data. DPT~\cite{DPT} leverages vision transformers using a scale- and shift-invariant trimmed loss. MiDas~\cite{midas} mixes multiple datasets with training objectives invariant to depth range and scale. Depth Anything~\cite{DepthAnything} and Depth Anything v2~\cite{DepthAnythingV2} train a depth foundation model using large-scale unlabeled data. Existing methods try to predict depth solely from images, which suffer from ambiguity like scale, object orientation, occlusion, and visual nuisance like viewpoints, illumination, appearance, texture, etc. In this paper, we investigate integrating language into diffusion-based monocular depth estimators to provide conditions of 3D scenes, to resolve the ambiguity and the nuisance variables in the pure visual algorithm.

\noindent
\textbf{Diffusion model for monocular depth estimation.} Denoising Diffusion Probabilistic Model (DDPM)~\cite{DDPM} learns to reverse a diffusion process that progressively degrades images with Gaussian noise so that they can draw samples from the data distribution by applying the reverse process to random noise. They have been applied to various tasks like text-to-image generation~\cite{StableDiffusion,ramesh2022hierarchical,zhu2024boundary,zhu2022discrete,wang2023diffusion}, super resolution~\cite{saharia2022image,li2022srdiff,gao2023implicit}, image inpainting~\cite{lugmayr2022repaint,corneanu2024latentpaint,yang2023uni}, 3D object generation~\cite{qian2023magic123,li20223ddesigner,liu2024one}, etc. Several approaches have explored the use of DDPM for metric depth estimation. DiffusionDepth~\cite{duan2023diffusiondepth} learns to denoise random depth distribution into a depth map with monocular visual conditions.  Similarly, DDP~\cite{ji2023ddp} encodes the input image and use diffusion to decodes it into a depth map. DepthGen~\cite{saxena2023monocular} extends a multi-task diffusion model for metric depth prediction. Its successor, DDVM~\cite{saxena2024surprising}, focuses on pretraining with synthetic and real datasets to improve depth estimation performance. Build upon the diffusion model, DepthFM~\cite{gui2024depthfm} uses flow matching to conduct a faster monocular depth estimation. 
In light of strong representation learned by pre-trained, text-to-image diffusion models, recent methods like VPD~\cite{zhao2023unleashing}, E2E-FT~\cite{garcia2024fine}, GeoWizard~\cite{fu2024geowizard}, Lotus~\cite{he2024lotus}, GenPercept~\cite{xu2024diffusion}, Jasmine~\cite{wang2025jasmine}, and Marigold~\cite{Marigold} utilize a pretrained Stable Diffusion model~\cite{StableDiffusion} as the backbone to predict depth through a denoising process. Specifically, Marigold~\cite{Marigold} concatenates input images with diffusion latents as conditions for the denoising U-Net to predict the noise to be removed, gradually denoising Gaussian noise into a depth map. While previous works have primarily focused on using images as the conditioning input to diffusion models for depth prediction, we investigate use language as an an additional condition to improve monocular depth estimation.

\noindent
\textbf{Language for monocular depth estimation.} Vision-Language models ~\cite{Dino,Blip-2,Blip,Dinov2,CLIP} acquire a comprehensive understanding of languages and images through pre-training under diverse datasets, thereby establishing a robust foundation for downstream tasks.~\cite{yang2024unitouch, yang2024neurobind, DSPoint, iQuery,PointCLIP,DepthCLIP,PointCLIP_v2,zeng2024wordepth,zeng2024rsa,ni2025efficient,tong2024metamorph,kim2024openvla,ni2025homer}.
For example, CLIP~\cite{CLIP} use contrastive learning on text-image pairs, which allows for various applications. These include few-shot image classification ~\cite{gao2021clip_cls,zhang2021tip_cls,VT-CLIP}, object detection ~\cite{rao2021denseclip_detect_seg,Detic}, image segmentation ~\cite{rao2021denseclip_detect_seg,zhou2021denseclip_seg}, and 3D perception \cite{DepthCLIPv2,PointCLIP,DepthCLIP,PointCLIP_v2,zhou2023uni3d,zhang2023clip,chen2023clip2scene}. Given their growing capabilities, some research~\cite{DepthCLIP_auty,DepthCLIPv2,DepthCLIP,VPD} has focused on applying vision-language models to monocular depth estimation. Specifically, 
DepthCLIP~\cite{DepthCLIP} utilizes CLIP's semantic depth response as an ordinal relationship alongside a depth projection method to conduct zero-shot monocular depth estimation. WorDepth \cite{zeng2024wordepth} learns the variational prior distribution of 3D scenes from text. RSA \cite{zeng2024rsa} predicts scale using language description to align relative depth to metric scale. Although language has been incorporated into several areas of 3D perception, its influence on diffusion-based monocular depth estimation has yet to be thoroughly investigated.


\section{Formulation}
\label{sec:method}

\textbf{Problem formulation.} Formally, given an RGB image $x : \Omega \subset \mathbb{R}^2 \rightarrow \mathbb{R}^3$, the objective of monocular depth estimation is to predict a dense depth map $y : \Omega \subset \mathbb{R}^2 \rightarrow \mathbb{R}_+$ using a parameterized function $h$, typically represented as a neural network, such that $y := h(x)$. We employ a supervised dataset $\mathcal{D} = \{x^{(m)}, \textbf{c}^{(m)}, y^{*(m)}\}_{m=1}^M$ comprised of $M$ samples, where $y^* : \Omega \subset \mathbb{R}^2 \rightarrow \mathbb{R}_+$ denotes the ground-truth depth map, and $\textbf{c}$ signifies the corresponding text caption of the image.

\noindent
\textbf{Diffusion formulation.} The pipeline to integrate text is shown in Figure~\ref{fig:pipeline}. Our way to use diffusion models for monocular depth estimation follows the formulation of ~\cite{duan2023diffusiondepth,ji2023ddp,saxena2023monocular,saxena2024surprising,zhao2023unleashing, Marigold}, adopting the formulation of Denoising Diffusion Probabilistic Model (DDPM)~\cite{DDPM} for the forward diffusion process, reverse diffusion process, training objective and inference mentioned below. A latent variable \(\mathbf{z}_t\) \( \in \mathbb{R}^{H' \times W' \times C'} \) is defined to represent a noisy version of the depth map in the latent space at diffusion timestep \( t \). 
The ground truch depth map \( y^* \) is encoded into a latent representation \( \mathbf{z_y} \) as supervision using a frozen VAE encoder \( E \): $\mathbf{z_y} = E(y^*)$. The diffusion latent after the whole denoising process \( \mathbf{z_0} \) is decoded back to the image space using the frozen VAE decoder \( D \): $\hat{y} = D(\mathbf{z}_0)$.

\begin{figure*}[t!]
  \centering
    \includegraphics[width=1.0\textwidth]{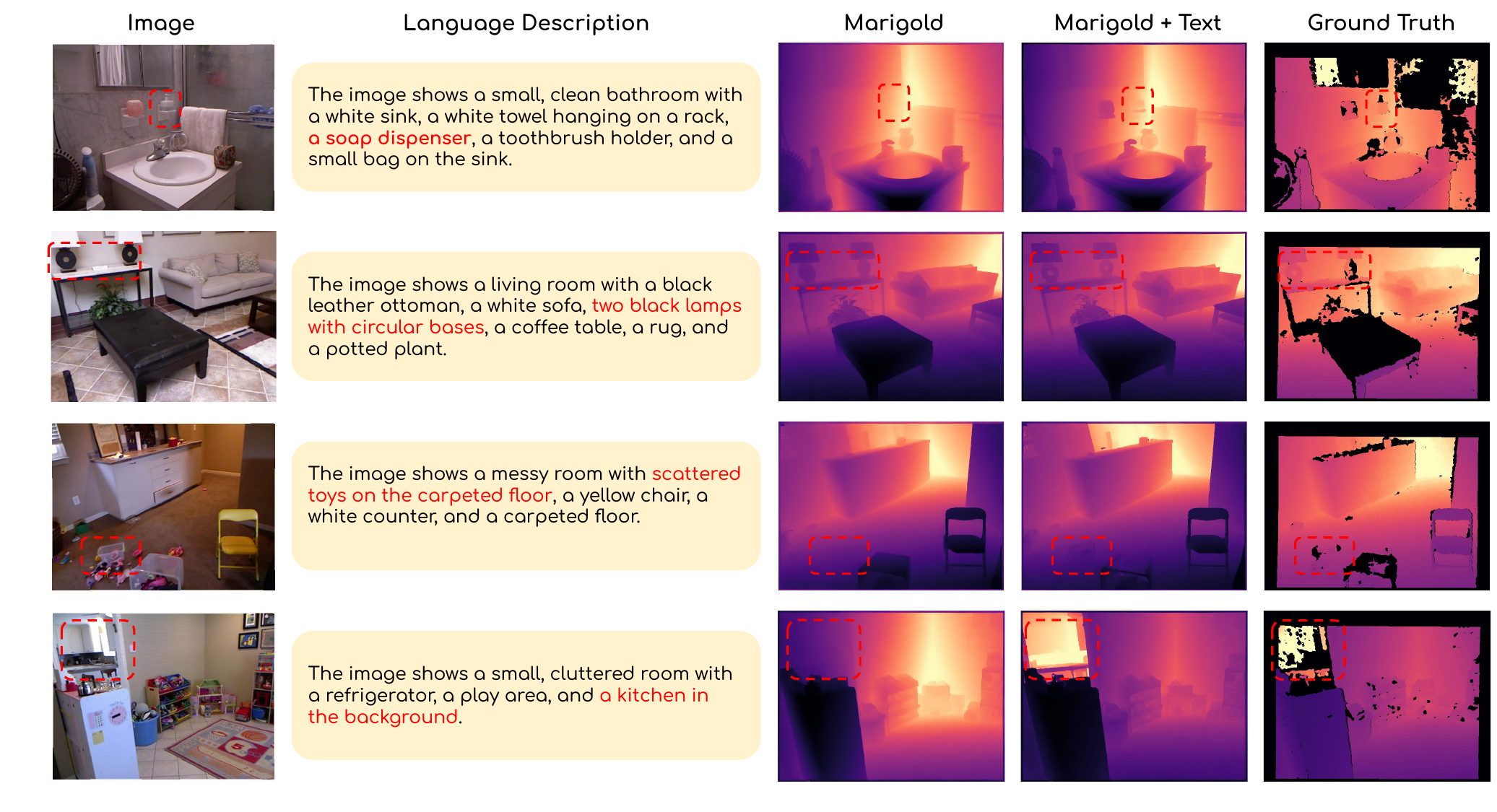}   \caption{\textbf{Visualization on NYUv2.} Compared to the Marigold baseline, integrating language demonstrates more accurate depth prediction for a given input image, particularly for instances specified in the language description (marked in red). This is achieved by providing additional language conditions for the semantic and geometric characteristics of specified objects. It's particularly beneficial for ambiguous or insignificant areas that are easily neglected by visual signals, like ``a soap dispenser'' in the first row, and ``two black lamps with circular bases'' in the second row.}
    \label{fig:vis_nyu}
    \vspace{-0.3cm}
\end{figure*}

\begin{figure*}[t!]
  \centering
    \includegraphics[width=1\textwidth]{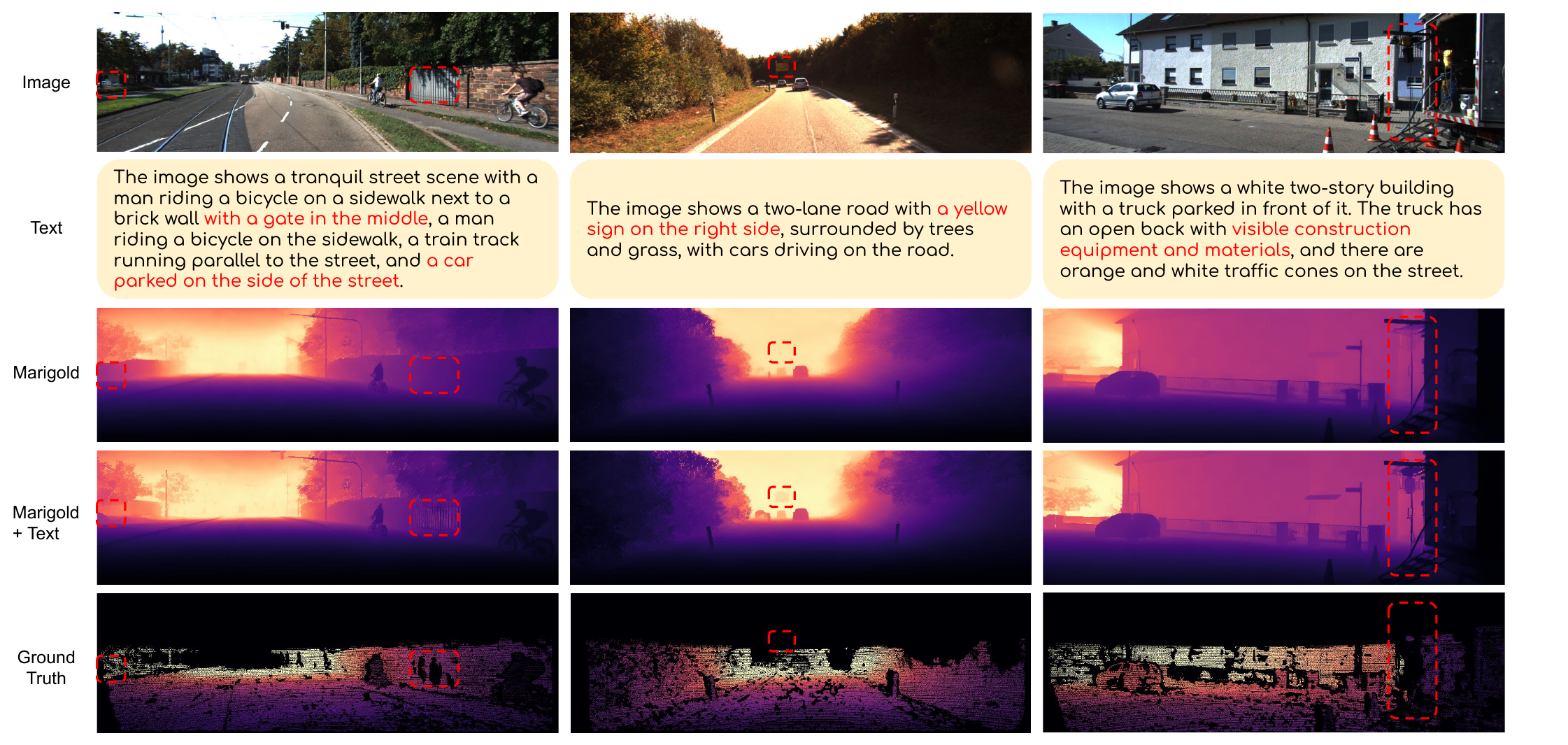}   \caption{\textbf{Visualization on KITTI.} Integrating language allows predicting better depth for described objects, even when parts of the object are almost invisible in the image (such as the parked car in the first column). Additional semantic and geometrical conditions are provided for the described ambiguous and insignificant regions, such as a sign at a distance, potentially enhancing the safety of self-driving systems that rely solely on vision sensors.}
    \label{fig:vis_kitti}
    \vspace{-0.3cm}
\end{figure*}

\noindent
\textbf{Forward diffusion process.} The forward process gradually adds noise to the latent depth feature in \( T \) steps. The process can be defined as:
\[
q(\mathbf{z}_{t} | \mathbf{z}_{t-1}) = \mathcal{N}(\mathbf{z}_{t}; \sqrt{1 - \beta_t} \mathbf{z}_{t-1}, \beta_t \mathbf{I}),
\]
where \( \beta_t \) is a scalar that defines the variance of the Gaussian noise added at each step \( t \) in the forward diffusion process. This parameter controls how much noise is added into the latent variable \( \mathbf{z}_t \) as it transitions from \( \mathbf{z}_{t-1} \) over time.

The forward process from \(y\) to \(\mathbf{z}_T\) can be written as:
\[
q(\mathbf{z}_{1:T} | y^*) = \prod_{t=1}^{T} q(\mathbf{z}_{t} | \mathbf{z}_{t-1}),
\]
where \( D(\mathbf{z}_0) = \hat{y} \).

\begin{table*}[t!]
\centering
\begin{adjustbox}{width=\textwidth}
\begin{tabular}{@{}lccccccccccccccc@{}}
\toprule
\multirow{2}{*}{\textbf{Method}} &\multirow{2}{*}{\textbf{Training Images}} & \multicolumn{2}{c}{\textbf{NYUv2}} & & \multicolumn{2}{c}{\textbf{KITTI}} & & \multicolumn{2}{c}{\textbf{ETH3D}} & & \multicolumn{2}{c}{\textbf{ScanNet}}& & \multicolumn{2}{c}{\textbf{DIODE}} \\
\cmidrule{3-4} \cmidrule{6-7} \cmidrule{9-10} \cmidrule{12-13} \cmidrule{15-16}
& & $\delta_1$↑ & AbsRel↓ & & $\delta_1$↑ & AbsRel↓ & & $\delta_1$↑ & AbsRel↓ & & $\delta_1$↑ & AbsRel↓ & & $\delta_1$↑ & AbsRel↓ \\
\midrule
GeoWizard~\cite{fu2024geowizard} &280K &96.3 &5.6 &&82.0 &14.4 &&\textbf{95.8} &\textbf{6.6} &&95.0 &6.4 &&79.2 &29.7\\
DPT~\cite{DPT} &1.4M & 90.3 & 9.8 & & 90.1 & 10.0 & & 94.6 & 7.8 & & 93.4 & 8.2 && 75.8 & 18.2\\
Omnidata~\cite{eftekhar2021omnidata} &12.2M & 94.5 & 7.4 & & 83.5 & 14.9 & & 77.8 & 16.6 & & 93.6 & 7.5 && 74.2 & 33.9\\
Depth Anything v2~\cite{DepthAnythingV2} &62.6M &97.9 &4.4 &&\textbf{94.8} &\textbf{7.5} && 86.2 &13.2 & &97.8 &\textbf{4.2} &&\textbf{95.4} &\textbf{6.5}\\
Depth Anything~\cite{DepthAnything} &63.5M &\textbf{98.1} &\textbf{4.3} & &94.7 &7.6 & &88.2 &12.7 & &\textbf{98.1} &4.3 &&95.2 &6.6\\

\midrule
Marigold~\cite{Marigold} &74K & \textbf{95.9} & 6.0 & &90.4 & 10.5 & & 95.1 & 7.1 & & 94.5 & 6.9  &&77.2 &31.0\\
Marigold$^*$ &74K & 95.7 & 6.1 & & 89.7 & 10.7 & & 95.4 & 6.9 & & 94.0 & 7.3 &&77.0 &31.3\\ [0.5ex]

Marigold + Text (Train Only) &74K  &95.5 &6.2				&&89.3 &10.9			&&95.0	&6.9		&&92.6 &8.1  &&77.3 &30.7\\

Marigold + Text (Infer Only) &74K &95.7 &6.1 &&89.8 &10.6 &&95.3 &6.9 &&93.8 &7.3 &&77.1 &31.0\\
\rowcolor{cyan!10}
\textbf{Marigold + Text (Train \& Infer)} &74K & \textbf{95.9} & \textbf{5.9} & & \textbf{90.6} & \textbf{10.4} & & \textbf{95.7} & \textbf{6.5} & & \textbf{94.9} & \textbf{6.7} &&\textbf{78.9} &\textbf{29.8}\\
\hdashline
\textcolor{gray}{Lotus-D~\cite{he2024lotus}} & \textcolor{gray}{59K} & \textcolor{gray}{97.2} & \textcolor{gray}{5.1} && \textcolor{gray}{93.1} & \textcolor{gray}{8.1} && \textcolor{gray}{97.0} & \textcolor{gray}{6.1} && \textcolor{gray}{96.5} & \textcolor{gray}{5.5} && \textcolor{gray}{73.8} & \textcolor{gray}{22.8}\\

Lotus-D$^*$ &59K &96.6 &5.6 &&92.2 &8.7 &&96.8 &6.1 &&96.0 &6.0 &&74.1 &22.2\\

Lotus-D + Text (Train Only) &59K &96.2 &5.8 &&91.7 &8.9 &&96.9 &6.1 &&95.1 &6.4 &&73.7 &24.0\\

Lotus-D + Text (Infer Only) &59K &96.7 &5.6 &&92.6 &8.5 &&96.7 &6.2 &&96.1 &5.8 &&74.1 &22.4\\

\rowcolor{cyan!10}
\textbf{Lotus-D + Text (Train \& Infer)} &59K &\textbf{96.8} &\textbf{5.4} &&\textbf{93.0} &\textbf{8.4} &&\textbf{97.0} &\textbf{6.0} &&\textbf{96.6} &\textbf{5.6} &&\textbf{74.2} &\textbf{22.0}\\

\hdashline
\textcolor{gray}{Lotus-G~\cite{he2024lotus}} & \textcolor{gray}{59K} & \textcolor{gray}{96.8} & \textcolor{gray}{5.4} && \textcolor{gray}{92.2} & \textcolor{gray}{8.5} && \textcolor{gray}{97.0} & \textcolor{gray}{5.9} && \textcolor{gray}{95.7} & \textcolor{gray}{5.9} && \textcolor{gray}{72.9} & \textcolor{gray}{22.9}\\

Lotus-G$^*$ &59K &95.2 &6.7 &&92.2 &8.9 &&95.7 &9.2 &&93.7 &7.6 &&71.7 &25.4\\

Lotus-G + Text (Train Only) &59K &95.8 &6.3 &&91.9 &9.2 &&95.6 &12.5 &&93.7 &7.3 &&71.9 &25.3\\

Lotus-G + Text (Infer Only) &59K &95.2 &6.7 &&92.2 &9.0 &&95.5 &9.6 &&94.1 &7.3 &&71.6 &28.7\\

\rowcolor{cyan!10}
\textbf{Lotus-G + Text (Train \& Infer)} &59K &\textbf{96.3} &\textbf{5.9} &&\textbf{92.8} &\textbf{8.6} &&\textbf{96.3} &\textbf{9.0} &&\textbf{95.3} &\textbf{6.4} &&\textbf{72.5} &\textbf{24.3}\\

\hdashline
\textcolor{gray}{E2E-FT (Stable Diffusion)~\cite{garcia2024fine}}&\textcolor{gray}{74K} & \textcolor{gray}{96.5} & \textcolor{gray}{5.4} && \textcolor{gray}{92.1} & \textcolor{gray}{9.6} && \textcolor{gray}{95.9} & \textcolor{gray}{6.4} && \textcolor{gray}{96.5} & \textcolor{gray}{5.8} && \textcolor{gray}{77.6} & \textcolor{gray}{30.3} \\

E2E-FT (Stable Diffusion)$^*$ &74K 
&95.4 &6.9 &&90.1 &10.5 &&94.1 &8.1 &&94.6 &7.7 &&76.4 &\textbf{31.0}\\

E2E-FT + Text (Train Only) &74K 
&95.6 &6.9 &&90.5 &10.3 &&93.3 &8.8 &&95.0 &7.5 &&76.5 &31.5\\

E2E-FT + Text (Infer Only) &74K 
&96.0	&6.4	&&90.8	&10.2	&&94.2	&8.2	&&\textbf{95.4} 	&\textbf{7.1}	&&76.4	&31.4\\

\rowcolor{cyan!10}
\textbf{E2E-FT + Text (Train \& Infer)} &74K 
&\textbf{96.3} &\textbf{6.2} &&\textbf{91.7} &\textbf{9.7} &&\textbf{94.7} &\textbf{7.8} &&95.0 &7.5 &&\textbf{77.0} &31.4\\

\bottomrule
\end{tabular}
\end{adjustbox}
\caption{\textbf{Quantitative comparison.} Integrating language generally outperforms the baselines. In some cases, integrating language only during training or only during inference can also lead to performance improvements. Metrics are reported as percentages. * indicates results that we re-trained and re-evaluated using their open-sourced code. Due to computational overhead, all models are evaluated without ensembling. We gray out results that were not reproducible with the released code and models. Unless otherwise specified, in this paper, integrating text means integrating text in both training and inference.}
\label{tab:eval_results}
\vspace{-0.3cm}
\end{table*}


\noindent
\textbf{Reverse diffusion process.} The goal of the reverse process is to denoise \(\mathbf{z}_T\) to recover \( \hat{y} \). The reverse process is defined as:
\[
p_{\theta}(\mathbf{z}_{0:T} | x, \mathbf{c}) = p(\mathbf{z}_T) \prod_{t=1}^{T} p_{\theta}(\mathbf{z}_{t-1} | \mathbf{z}_{t}, x, \mathbf{c}),
\]
where \( p(\mathbf{z}_T) \) is assumed to be a standard Gaussian distribution \( \mathcal{N}(\mathbf{z}_T; \mathbf{0}, \mathbf{I}) \), and the denoising step \( p_{\theta}(\mathbf{z}_{t-1} | \mathbf{z}_{t}, x, \mathbf{c} )\) is parameterized as:
\[
p_{\theta}(\mathbf{z}_{t-1} | \mathbf{z}_{t}) = \mathcal{N}(\mathbf{z}_{t-1}; \mu_{\theta}(\mathbf{z}_{t}, t, x, \mathbf{c}), \Sigma_{\theta}(\mathbf{z}_{t}, t, x, \mathbf{c})),
\]
where \( \mu_{\theta} \) and \( \Sigma_{\theta} \) is the diffusion model, that predicts the mean and variance conditioned on \( \mathbf{z}_{t} \), \( t \), the input image \( x \) and the corresponding text \( \mathbf{c} \). The diffusion model is initialized by the denoising U-Net from Stable Diffusion v2~\cite{StableDiffusion}. Specifically, given a text caption $\textbf{c} = \{ c_1, c_2, ... \}$, we first encode it through a frozen CLIP text encoder, then feed it into the diffusion model. Given an image $x$, we encode it using the same VAE encoder $E(x)$ that encoded the depth map, concatenated with the depth latent $\mathbf{z}_t$, then feed it into the diffusion model.

\noindent
\textbf{Training objective.} The training objective is derived using the reparameterization trick and variational bounds, which involves predicting the noise added at each step. The loss function is:
\[
\mathcal{L}(\theta) = \mathbb{E}_{y, \epsilon, t} \left[ \| \epsilon - \epsilon_{\theta}(\mathbf{z}_t, t, x, \mathbf{c}) \|^2 \right],
\]
where:
\( \epsilon \sim \mathcal{N}(\mathbf{0}, \mathbf{I}) \) is the noise sampled during the forward process, and
\( \mathbf{z}_t = \sqrt{\bar{\alpha}_t} \mathbf{z_y} + \sqrt{1 - \bar{\alpha}_t} \epsilon \) is the noisy depth map at step \( t \), where \( \bar{\alpha}_t = \prod_{s=1}^{t} (1 - \beta_s) \).

\noindent
\textbf{Inference.} The inference phase begins by sampling a purely noisy latent variable \( \mathbf{z}_T \) from a standard Gaussian distribution, \( \mathbf{z}_T \sim \mathcal{N}(\mathbf{0}, \mathbf{I}) \). This initial noisy sample represents the starting point of the reverse diffusion process. The goal is to progressively refine this sample through a series of denoising steps until the final depth latent \( \mathbf{z}_0 \) is obtained. Each denoising step is to predict the noise component \( \epsilon_{\theta}(\mathbf{z}_t, t, x, \mathbf{c}) \) that needs to be removed from the latent variable at time step \( t \). The prediction is made by the diffusion model conditioned on the noisy latent \( \mathbf{z}_t \), the current timestep \( t \), the input image \( x \), and the corresponding text description \( \mathbf{c} \). The iterative denoising process follows the reverse transition:
\[
\mathbf{z}_{t-1} = \frac{1}{\sqrt{\alpha_t}} \left( \mathbf{z}_t - \frac{1 - \alpha_t}{\sqrt{1 - \bar{\alpha}_t}} \epsilon_{\theta}(\mathbf{z}_t, t, x, \mathbf{c}) \right),
\]
where \( \alpha_t \) and \( \bar{\alpha}_t \) are derived from the noise schedule \( \beta_t \). This reverse diffusion continues until \( t = 0 \), at which point the fully denoised latent \( \mathbf{z}_0 \) is reached. The predicted depth map \( \hat{y} \) is obtained by decoding \( \mathbf{z}_0 \) back into the image space using the pre-trained and frozen VAE decoder:\(\hat{y} = D(\mathbf{z}_0). \)

Unless otherwise specified,``integrating text'' in this paper is in both diffusion training and inference.




\begin{figure*}[t]
  \centering
  \vspace{0.3cm}
\includegraphics[width=0.95\textwidth]{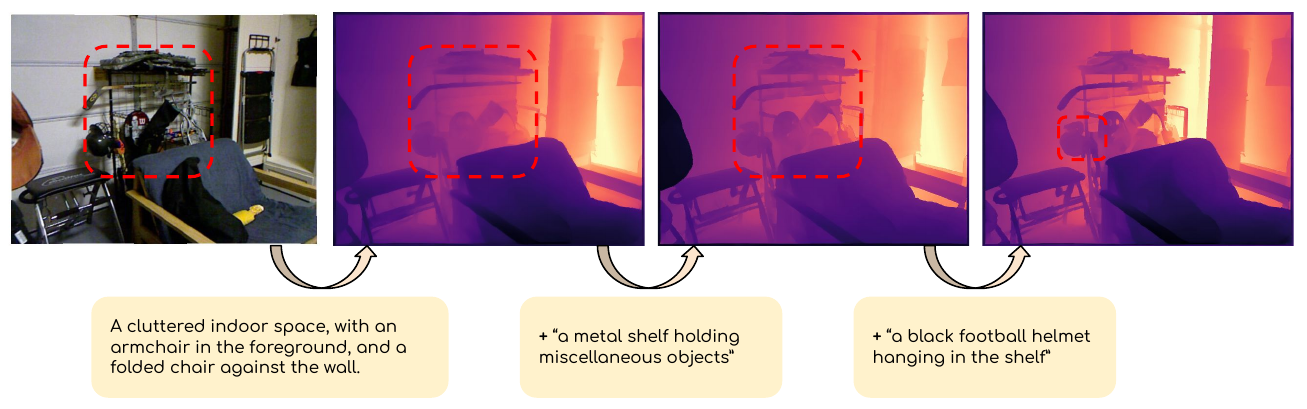}
    \caption{
        \textbf{Iterative depth refinement with language}. Each time, we append the newly added sentence to the end of the original sentence and repeat the inference process. Is shows that by providing more details in the language, depth prediction can be iteratively refined, especially for specified regions or objects. 
    }
    \label{fig:vis_control}
    \vspace{-0.3cm}
\end{figure*}

\section{Experiments}
\label{sec:experiments}

\textbf{Training datasets.} The training scheme follows the same ones proposed in Marigold~\cite{Marigold}, Lotus~\cite{he2024lotus}, and E2E-FT~\cite{garcia2024fine}. All of them are trained on two synthetic datasets that cover both indoor and outdoor scenes. The first dataset utilized is HyperSim~\cite{roberts2021hypersim}, a photorealistic compilation of 461 indoor scenes. Marigold and E2E-FT approximately select 54,000 samples, and Lotus filters out incomplete samples to keep approximately 39,000 samples. The second dataset is Virtual KITTI~\cite{VirtualKITTI, gaidon2016virtual}, which comprises synthetic street scenes across five distinct settings, with varied conditions such as different weather patterns and camera perspectives. The Virtual KITTI 2~\cite{VirtualKITTI} version is used, and for all Marigold, Lotus, and E2E-FT, four scenes are selected with totaling approximately 20,000 samples.


\noindent
\textbf{Evaluation datasets.} Trained models are evaluated using 5 real-world datasets that were not part of their training data: NYUv2~\cite{NYU-Depth-V2}, ScanNet~\cite{ScanNet}, KITTI dataset~\cite{KITTI, KITTI_1, Eigen-Split}, ETH3D~\cite{ETH3D}, and DIODE~\cite{DIODE}, following Marigold~\cite{Marigold}, Lotus~\cite{he2024lotus}, and E2E-FT~\cite{garcia2024fine}. 
Following the affine-invariant depth evaluation protocol~\cite{zeng2024rsa,DPT,DepthAnything,Marigold,midas}, we evaluate first-order threshold accuracy $\delta_{1}$ and mean absolute relative error $Abs Rel$. Details of the evaluation data and metric are provided in the Supplementary Material.


\noindent
\textbf{Obtain language description.} To implement our method, we require human-provided text descriptions for each image. Since standard benchmarks do not include such descriptions, we propose using off-the-shelf models to generate text descriptions to simulate those a human would provide. For this purpose, we use Vision Language Models to generate text. Specifically, we use LLaVA v1.6~\cite{liu2023improved} for Marigold, and InternVL3-8B~\cite{zhu2025internvl3} for Lotus and E2E-FT. Each training and testing image is generated with one text descriptions. Details of obtaining language descriptions can be found in the Supplementary Material.



\noindent
\textbf{Implementation details.} We follow the setup of original Marigold~\cite{Marigold}, Lotus~\cite{he2024lotus}, and E2E-FT~\cite{garcia2024fine}. Details are provided in the supplementary materials.


\noindent
\textbf{Quantitative comparison.} Shown in Table~\ref{tab:eval_results}, after integrating text, the accuracy of diffusion-based monocular depth estimators are generally improved over the baseline.  We also find that in some cases, integrating language only during training or only during inference can lead to performance improvements. It shows that language conditions can potentially regularize the diffusion learning process and may also guide depth prediction in a zero-shot manner.



\noindent
\textbf{Qualitative comparison.} Besides the overall quantitative improvement, we find that language can particularly benefit specified insignificant or ambiguous regions. The visualization for NYUv2~\cite{NYU-Depth-V2} is shown in Figure~\ref{fig:vis_nyu}, where integrating text achieves more accurate depth prediction for a given input image compared to the baseline, especially for instances highlighted in the language description. For example, in the first row, the soap dispenser is barely distinguishable in the input image due to its color and texture blending with the background. Vision-based depth estimators like vanilla Marigold struggle to identify it, resulting in a depth prediction that matches the background. In contrast, the language description explicitly includes the term ``a soap dispenser,'' providing the model with extra conditions that indicate the existence of a soap dispenser and what it should look like, improving its depth estimation. Similarly, as shown in Figure~\ref{fig:vis_kitti}, we visualize results for the KITTI~\cite{KITTI} dataset, where objects such as a parked car or a sign are barely discernible from the visual signal alone. As shown in Figure~\ref{fig:vis_control}, by incorporating more detailed language descriptions, depth predictions are enhanced for the specified regions or objects.

\begin{figure}[t]
  \centering
  \vspace{0.3cm}
\includegraphics[width=0.48\textwidth]{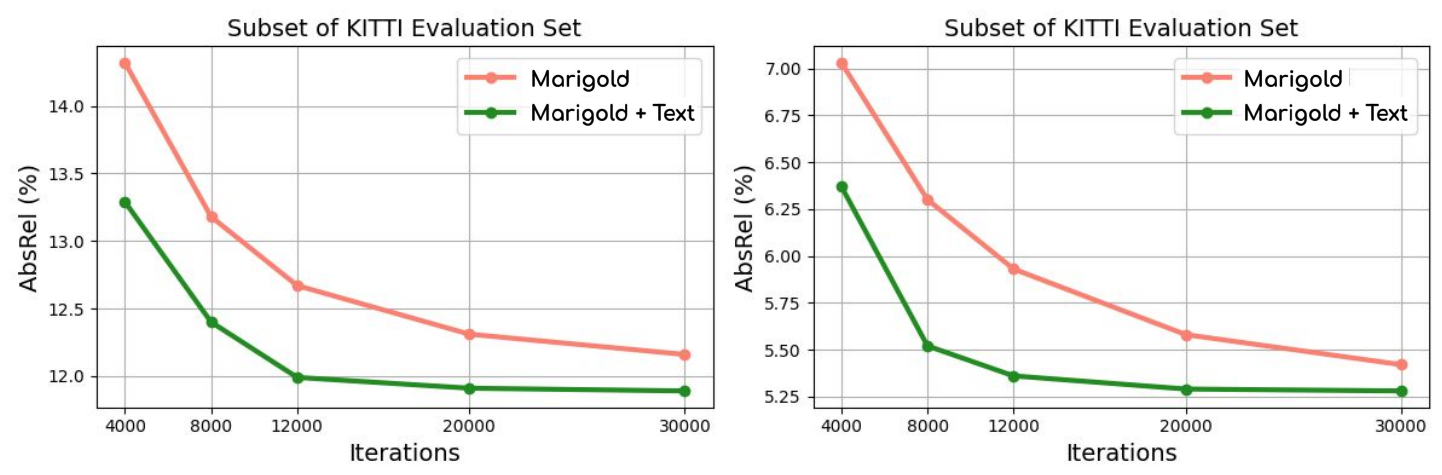}
    \caption{
        \textbf{Convergence speed comparison}. Integrating language converges faster during training compared with the Marigold baseline.
    }
    \label{fig:training_converge_speed}
    \vspace{-0.3cm}
\end{figure}

\noindent
\textbf{Faster training convergence.} One of the key advantages of integrating text is its faster convergence compared to Marigold. As shown in Figure~\ref{fig:training_converge_speed} here, the improved convergence can be attributed to the integration of language in the diffusion model, which provides additional semantic and geometric constraints that help to learn the diffusion process. Due to computation overhead, following the same configuration of Marigold, we use a subset of KITTI and NYUv2 for training-time evaluation.

\begin{figure}
    \centering
    \includegraphics[width=\linewidth]{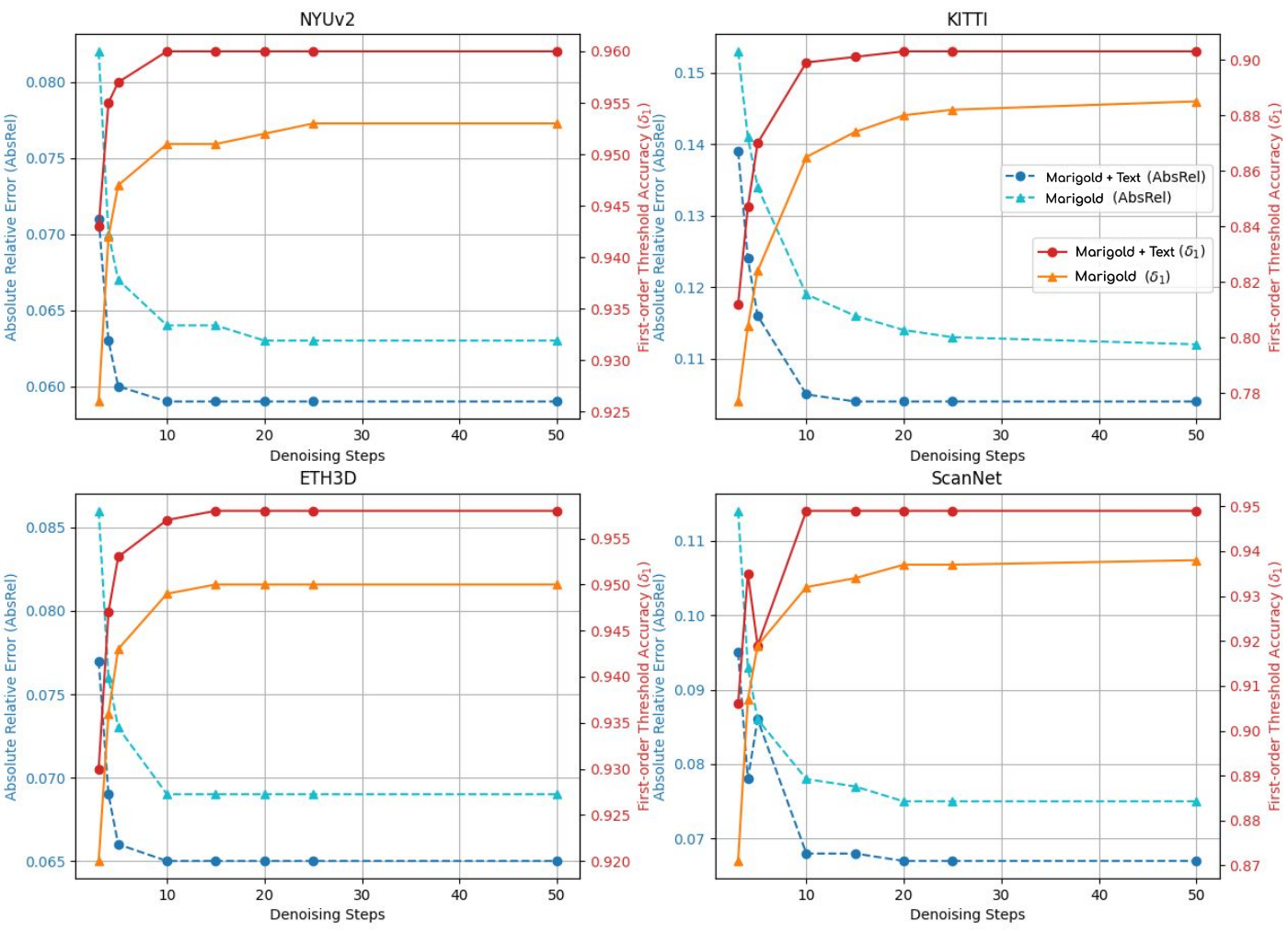}
    \caption{\textbf{Performance under different denoising steps.} Integrating language consistently outperforms the Marigold baseline across various denoising steps, with a faster convergence.}
    \label{fig:denoising_steps}
    \vspace{-0.5cm}
\end{figure}

\noindent
\textbf{Fewer denoising steps.}
As demonstrated in Figure~\ref{fig:denoising_steps}, and detailed in Table~\ref{tab:denoising_steps} in the supplementary material, we evaluate Marigold with integrating text and the baseline using different denoising steps during inference. We find that integrating text consistently outperforms the baseline across different denoising steps. Remarkably, integrating converges in just 10 steps, while the baseline takes 25 steps to achieve convergence. One explanation is that language provides additional semantic and geometric constraints that speed up the diffusion process.

\noindent\textbf{Better perception in small areas.}
To demonstrate a better perception for small objects and areas mentioned in text, as shown in Figure~\ref{fig:vis_nyu}. We use MaskDINO to obtain panoptic segmentation masks for all images in NYUv2 test set, then evaluate the performance for masks that occupied less than 5\%, 10\%, and 20\% of an image. Table~\ref{tab:small_areas} shows that small areas are harder to perceive compared with large areas, and integrating text obtains better improvement over small areas, as language provides a condition for geometric and semantic properties that guide the diffusion model to better perceive small areas and tiny structures. 

\begin{table}[t]
\centering
\begin{adjustbox}{width=\linewidth}
\begin{tabular}{@{}lccccccccccc@{}}
\toprule
\multirow{2}{*}{NYUv2} & \multicolumn{2}{c}{Standard} & & \multicolumn{2}{c}{Small Area (5\%)}  & & \multicolumn{2}{c}{Small Area (10\%)} & & \multicolumn{2}{c}{Small Area (20\%)}\\
\cmidrule{2-3} \cmidrule{5-6} \cmidrule{8-9} \cmidrule{11-12} 
& $\delta_1$↑ & AbsRel↓ & & $\delta_1$↑ & AbsRel↓ && $\delta_1$↑ & AbsRel↓ & & $\delta_1$↑ & AbsRel↓\\
\midrule
Marigold &95.7 &6.1 &&91.7 &9.0 &&92.2 &8.4 &&93.9 &7.3\\

Marigold + Text  &\textbf{95.9} &\textbf{5.9} &&\textbf{92.8} &\textbf{8.3} &&\textbf{93.1} &\textbf{7.9} &&\textbf{94.6} &\textbf{6.8}\\

\bottomrule
\end{tabular}
\end{adjustbox}
\caption{\textbf{Better perception in small areas.} We evaluate the performance for panoptic segmentation masks that occupied less than 5\%, 10\%, and 20\% of an image. Integrating language results in better depth estimation in small areas.}
\label{tab:small_areas}
\vspace{-4mm}
\end{table}

\begin{figure}[t!]
    \includegraphics[width=\linewidth]{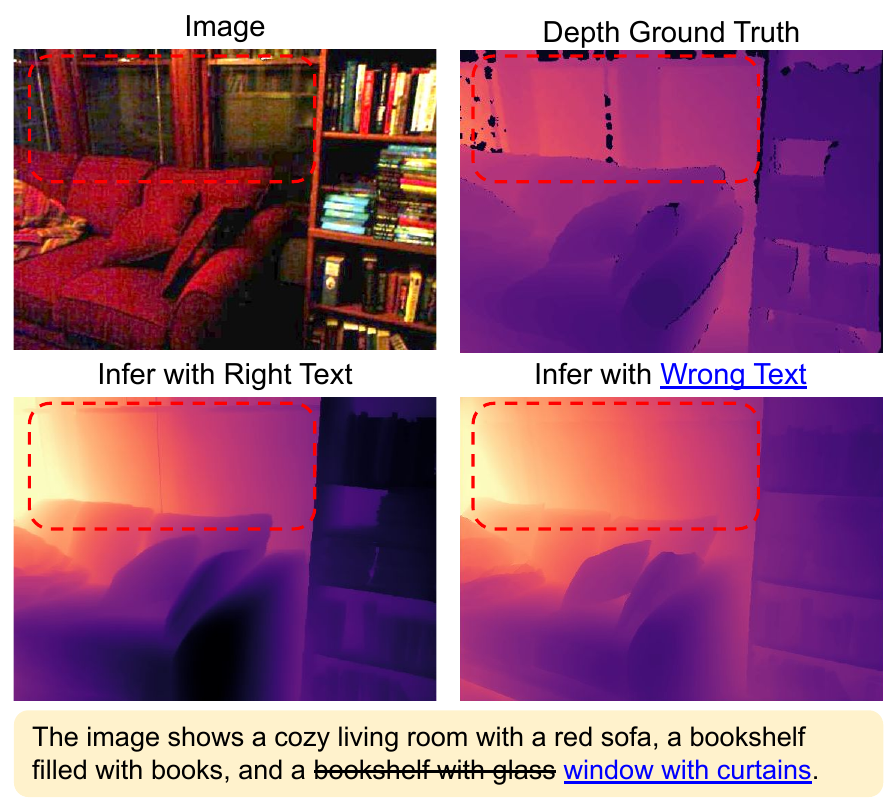}
    \vspace{-6mm}
    \caption{\textbf{Incorrect text misguides models.}  After replacing the correct description \textit{``a bookshelf with glass''} with a wrong description \textit{``a window with curtains''}, the model failed to perceive the structure of the bookshelf behind the glass.}
    \label{fig:failure_case}
    \vspace{-5mm}
\end{figure}

\section{Discussion}
\label{sec:discussion}

\noindent
\textbf{Conclusion.} 
We investigate the influence of integrating language into diffusion-based depth estimators and find that it enhances monocular depth estimation. Our explanation is that language provides an additional condition (learned during the text-to-image pre-training of diffusion models) associated with plausible 3D scenes, thus reducing the solution space for depth estimation. 




\noindent
\textbf{Limitation.} One primary limitation is its dependence on the accuracy and detail of the provided language descriptions. Shown in Figure~\ref{fig:failure_case}, ambiguous or misleading text inputs from human users can result in suboptimal depth predictions and potentially compromise the model's performance. Although human-provided descriptions are typically more accurate than automatically generated ones, they can still be incomplete or contain subtle ambiguities. A potential solution is to incorporate a language-robustness module, such as uncertainty estimation or consistency-based filtering, to correct or filter out incorrect descriptions.



{
    \small
    \bibliographystyle{ieeenat_fullname}
    \bibliography{main}
}

\input{X_suppl}
\end{document}

%% file: X_suppl.tex
\clearpage
\setcounter{page}{1}

\maketitlesupplementary

\section*{A. Dataset Details}
We train our model on two synthetic datasets, Hypersim~\cite{roberts2021hypersim} and Virtual Kitti~\cite{VirtualKITTI}, and conduct zero-shot evaluations on four additional real-world datasets that were not part of its training data, NYUv2~\cite{NYU-Depth-V2}, KITTI~\cite{KITTI}, ScanNet~\cite{ScanNet}, and ETH3D~\cite{ETH3D}. Details of each dataset are provided below.

\subsection*{A.1. Training Datasets}

\noindent
\textbf{Hypersim}~\cite{roberts2021hypersim} is a photorealistic synthetic dataset designed for comprehensive indoor scene understanding, and is introduced since obtaining per-pixel ground truth labels from real images is often challenging or impossible for many essential scene understanding tasks. This dataset is created using a vast collection of synthetic scenes developed by professional artists, resulting in 77,400 images across 461 indoor scenes with detailed per-pixel annotations and corresponding ground truth geometry. HyperSim is built exclusively using publicly accessible 3D assets. It includes complete scene geometry, material properties, and lighting information for each scene. Also, it provides dense per-pixel semantic instance segmentations and comprehensive camera details for each image. Further, it decomposes each image into diffuse reflectance, diffuse illumination, and a non-diffuse residual component that captures view-dependent lighting effects. In terms of training split, for Marigold and E2E-FT, as mentioned in the Experiments Section, we utilize the official dataset split to select approximately 54,000 samples from 365 scenes, and the RGB images and depth maps are resized to a resolution of 480 × 640 pixels. For Lotus, approximately 39,000 samples are selected, and the RGB images and depth maps are resized to a resolution of 576 × 768 pixels. The original distance measurements, defined relative to the focal point, are transformed into standard depth values relative to the focal plane.

\noindent
\textbf{Virtual Kitti}~\cite{VirtualKITTI,gaidon2016virtual} 
Virtual KITTI is a photorealistic synthetic video dataset created for training and evaluating computer vision models on various video understanding tasks, including object detection, multi-object tracking, scene-level and instance-level semantic segmentation, optical flow, and depth estimation. The dataset comprises 50 high-resolution monocular videos (a total of 21,260 frames) generated from five distinct virtual urban environments, each presented under varying imaging and weather conditions. These virtual scenes were developed using the Unity game engine and an innovative real-to-virtual cloning technique. The synthetic videos come with precise, fully automatic annotations for 2D and 3D multi-object tracking, as well as per-pixel category, instance, flow, and depth labels. We use its upgraded version, Virtual KITTI 2~\cite{VirtualKITTI}, which consists of the same five sequence clones as Virtual KITTI, with increased photorealism. It takes advantage of recent advancements in lighting and post-processing within the game engine, making the variations in the virtual sequences more closely mimic real-world changes in conditions. For training, we choose four scenes, comprising around 20,000 samples, and crop the images to match the resolution of the KITTI benchmark~\cite{KITTI}. The maximum depth is capped at 80 meters. For all models, the resolution is set to 352 × 1216.

\subsection*{A.2. Evaluation Datasets}
\noindent
\textbf{NYUv2}~\cite{NYU-Depth-V2} dataset comprises 24,231 synchronized RGB images and depth maps at a resolution of $640 \times 480$, representing various indoor scenes such as homes, offices, and commercial spaces, captured using a Microsoft Kinect. The standard split includes 249 training scenes and 215 test scenes. For our experiments, we use the official test set. Consistent with prior works~\cite{bhat2021adabins, yuan2022neural, zeng2024wordepth, zeng2024rsa, Marigold}, we exclude samples without valid ground truth, resulting in 654 valid images for evaluation. We perform evaluation on NYUv2 over a depth range spanning from $1 \times 10^{-3}$ to 10 meters.

\noindent
\textbf{KITTI}~\cite{KITTI, KITTI_1} contains 61 driving scenes with research in autonomous driving and computer vision. It contains calibrated RGB images with synchronized point clouds from Velodyne lidar, inertial, GPS information, etc. Following prior works~\cite{bhat2021adabins, yuan2022neural, zeng2024wordepth, zeng2024rsa, Marigold}, we used Eigen split~\cite{Eigen-Split}. It consists of 652 testing images after filtering out images without valid ground truth. We follow the evaluation protocol of~\cite{Eigen-Split-Eval} for our experiments.

\noindent
\textbf{ScanNet}~\cite{ScanNet} is an extensive RGB-D video dataset containing 2.5 million views in more than 1500 scans, annotated with 3D camera poses, surface reconstructions, and instance-level semantic segmentations. Data was collected using an RGB-D capture system (with a Kinect sensor) that includes automated surface reconstruction and crowdsourced semantic annotation. We use the same evaluation configuration of Marigold~\cite{Marigold}, where 800 images are randomly selected from the 312 official validation scenes for evaluation.

\noindent
\textbf{ETH3D}~\cite{ETH3D} is a multi-view stereo and 3D reconstruction benchmark encompassing a diverse range of indoor and outdoor scenes. High-precision laser scanning was used to obtain the ground truth geometry. Images were captured using both a DSLR camera and a synchronized multi-camera rig with varying fields-of-view. For evaluation, following Marigold~\cite{Marigold}, we use all 454 samples that include ground truth depth maps.

\section*{B. Implementation Details}

\subsection*{B.1. Visualization}
When visualizing the ground truth depth map, we apply the same affine transformation we used in training. As described in the Experiments Section, we apply a linear normalization ensuring that the depth values primarily fall within the range $[-1, 1]$. The affine transformation for normalization is defined as:
\begin{equation}
\tilde{y*} = \left( \frac{y^* - y_2}{y_2 - y_{98}} - 0.5 \right) \times 2,
\end{equation}
where $y_{2}$ and $y_{98}$ represent the 2\% and 98\% percentiles of the depth maps, respectively. 
Then, we apply min-max normalization to both the ground truth and predicted depth maps, scaling them to integer values within the range [0, 255]. These normalized depth maps are then visualized using the OpenCV MAGMA colormap.

It is important to note that, unlike Marigold~\cite{Marigold}, which applies linear fitting to the ground truth for error correction, we do not use this approach. While linear fitting can adjust predictions to more closely align with the ground truth, it does not accurately reflect the true distribution of the depth map predictions or provide a clear assessment of prediction quality. Instead, we enhance visualization by applying the same training normalization to the zero-shot evaluation dataset, avoiding linear fitting and its error correction effects, resulting in more authentic and insightful visualization outcomes.

\subsection*{B.2. Training Details}
\textbf{Marigold.} For Marigold~\cite{Marigold}, we implemented our method using PyTorch, employing Stable Diffusion v2~\cite{StableDiffusion} as the backbone and maintaining the original pre-training configuration with the $v$-objective~\cite{salimans2022progressive}. The training process utilized the DDPM noise scheduler~\cite{DDPM} with 1,000 diffusion steps, while at inference time, the DDIM scheduler~\cite{DDIM} was employed with 50 sampling steps for faster results. Our training setup spanned 30,000 iterations, with an effective batch size of 32 achieved through gradient accumulation over 16 steps (with a per-step batch size of 2) to fit on a single Nvidia RTX 3090 GPU. We used the Adam optimizer with a learning rate set at $3 \cdot 10^{-5}$ and included random horizontal flipping with a probability of 0.5 as data augmentation. For depth normalization, we employed a scale and shift-invariant method with clipping enabled and set the normalization range between -1.0 and 1.0, using a 0.02 min-max quantile to maintain robustness. This normalization strategy was applied during training and zero-shot evaluations for consistency. The training noise scheduler initialized from the pre-trained Stable Diffusion v2~\cite{StableDiffusion} model maintained a noise strength of 0.9 and incorporated an annealed strategy to progressively reduce noise levels. We saved checkpoints every 50 iterations, with backup, validation, and visualization checkpoints set at intervals of 2,000 iterations. The training process typically converged after approximately 20,000 iterations, though we extended training to 30,000 iterations for thorough coverage. We used mean squared error (MSE) as the loss function, with reduction set to ``mean'' for averaged loss calculation. A customized iteration-wise exponential scheduler is applied which adjusts the learning rate iteratively using an exponential decay function. It decays the learning rate to 1\% of its initial value over 25,000 iterations with a warmup phase of 100 steps. For text generation, generating a single caption for an image using LLaVA v1.6 on an RTX 3090 takes approximately 3.6 seconds, and we generate 10 captions for each image.  For training, we generate 740,000 captions, using 740 GPU hours on an RTX 3090. 

\noindent
\textbf{Lotus.}
For Lotus, we implemented our method using PyTorch, employing Stable Diffusion v2~\cite{StableDiffusion} as the backbone and maintaining the original pre-training configuration. The training process utilized the DDPM noise scheduler~\cite{DDPM} with 1,000 diffusion steps (inherited from the pre-trained Stable Diffusion v2 configuration), while at inference time, the DDIM scheduler~\cite{DDIM} was employed with 1 sampling step. Our training setup spanned 20,000 iterations, with an effective batch size of 36 achieved through gradient accumulation over 3 steps (with a per-step batch size of 4) across 3 Nvidia RTX 3090 GPUs. We used the 8-bit Adam optimizer with a learning rate set at $3 \cdot 10^{-5}$ and a constant learning rate scheduler without warmup. We included random horizontal flipping with a probability of 0.5 as data augmentation. For depth normalization, we employed a truncated disparity method, which was applied during training and zero-shot evaluations for consistency. We saved checkpoints every 500 iterations for Lotus-D and every 1,000 iterations for Lotus-G, with validation checkpoints set at the same intervals. The training process typically converged after approximately 20,000 iterations. The training utilized a mixed dataset strategy, combining Hypersim~\cite{roberts2021hypersim} at a resolution of $576 \times 576$ (sampled with 90\% probability) and VKITTI~\cite{VirtualKITTI} at a resolution of $375 \times 375$ (sampled with 10\% probability). For text generation, we used InternVL3-8B~\cite{zhu2025internvl3} to generate a single caption for each image. The generation process employed a specialized prompt focusing on depth estimation attributes (camera factors, scene properties, relative distances, object types, scales, illuminations, texture, visual features, occlusions, and boundaries), with a maximum token limit of 77 tokens per caption. 

\noindent
\textbf{E2E-FT.} For E2E-FT~\cite{garcia2024fine}, we implemented our method using PyTorch, employing Stable Diffusion v2~\cite{StableDiffusion} as the backbone and maintaining the original pre-training configuration with the $v$-objective~\cite{salimans2022progressive}. The training process utilized the DDPM noise scheduler~\cite{DDPM} with 1,000 diffusion steps (inherited from the pre-trained Stable Diffusion v2 configuration), while at inference time, the DDIM scheduler~\cite{DDIM} was employed with 1 sampling step. Our training setup spanned 20,000 iterations, with an effective batch size of 32 achieved through gradient accumulation over 16 steps (with a per-step batch size of 1) across 2 Nvidia RTX 3090 GPUs. We used the Adam optimizer with a learning rate set at $3 \cdot 10^{-5}$ and a customized iteration-wise exponential learning rate scheduler that decays the learning rate to 1\% of its initial value over 20,000 iterations with a warmup phase of 100 steps. We included random horizontal flipping with a probability of 0.5 as data augmentation. We employed mixed precision training with bfloat16 (bf16) for improved efficiency. For depth normalization, we employed a quantile-based method with clipping enabled, using the 0.02 and 0.98 quantiles to remove outliers and then normalizing the depth values to the range [-1.0, 1.0]. This normalization strategy was applied during training and zero-shot evaluations for consistency. The training utilized a mixed dataset strategy, combining Hypersim~\cite{roberts2021hypersim} (sampled with 90\% probability) and VKITTI~\cite{VirtualKITTI} (sampled with 10\% probability). For text generation, we used InternVL3-8B~\cite{zhu2025internvl3} to generate a single caption for each image. The generation process employed a specialized prompt focusing on depth estimation attributes (camera factors, scene properties, relative distances, object types, scales, illuminations, texture, visual features, occlusions, and boundaries), with a maximum token limit of 77 tokens per caption.

\subsection*{B.3. Evaluation metric}
Following the affine-invariant depth evaluation protocol~\cite{zeng2024rsa,DPT,DepthAnything,Marigold,midas}, for each image and the predicted relative depth $y$, we fit a pair of scalars denoting the scale and shift parameters of the transformation: $(\hat{\alpha}, \hat{\beta}) = g_\psi(y,y^*) \in \mathbb{R}^{2}$. The metric depth prediction is obtained by $\hat{y} =\hat{\alpha} \cdot y + \hat{\beta}$ such that:
\begin{equation}
 \psi^* = \arg\min_\psi \frac{1}{|{M}|} \sum_{(i,j) \in \Omega} M(i,j) | \hat{y}(i,j) - y(i,j)|
\end{equation}
where $\hat{y} = \hat{\alpha} \cdot y + \hat{\beta}$ denotes the predicted metric-scale depth aligned from relative depth $y$, $(i,j) \in \Omega$ denotes an image coordinate, and $M : \Omega \mapsto \{0, 1\}$ denotes a binary mask indicating valid coordinates in the ground truth depth $y^*$ with values greater than zero. Then, we follow~\cite{chang2021transformer, Va-depthnet,DepthCLIP,zeng2024wordepth,zeng2024rsa} to evaluate using first-order threshold accuracy, calculated as:
\begin{equation}
\delta_{1} = \% \text { of }  y(i,j) \text { s.t. } \max (\frac{{y}(i,j)
}{y^*(i,j)}, \frac{y^*(i,j)}{{y}(i,j)}) < 1.25
\end{equation}
and mean absolute relative error, calculated 
as: 
\begin{equation}
Abs Rel = \frac{1}{|{M}|}\sum_{(i,j)\in \Omega}  \frac{|y^*(i,j) - {y}(i,j) |}{y^*(i,j)}
\end{equation}

\section*{C. Additional Experiments}

\subsection*{C.1. Additional visualizations}

\begin{figure*}[t!]
  \centering
    \includegraphics[width=1.0\textwidth]{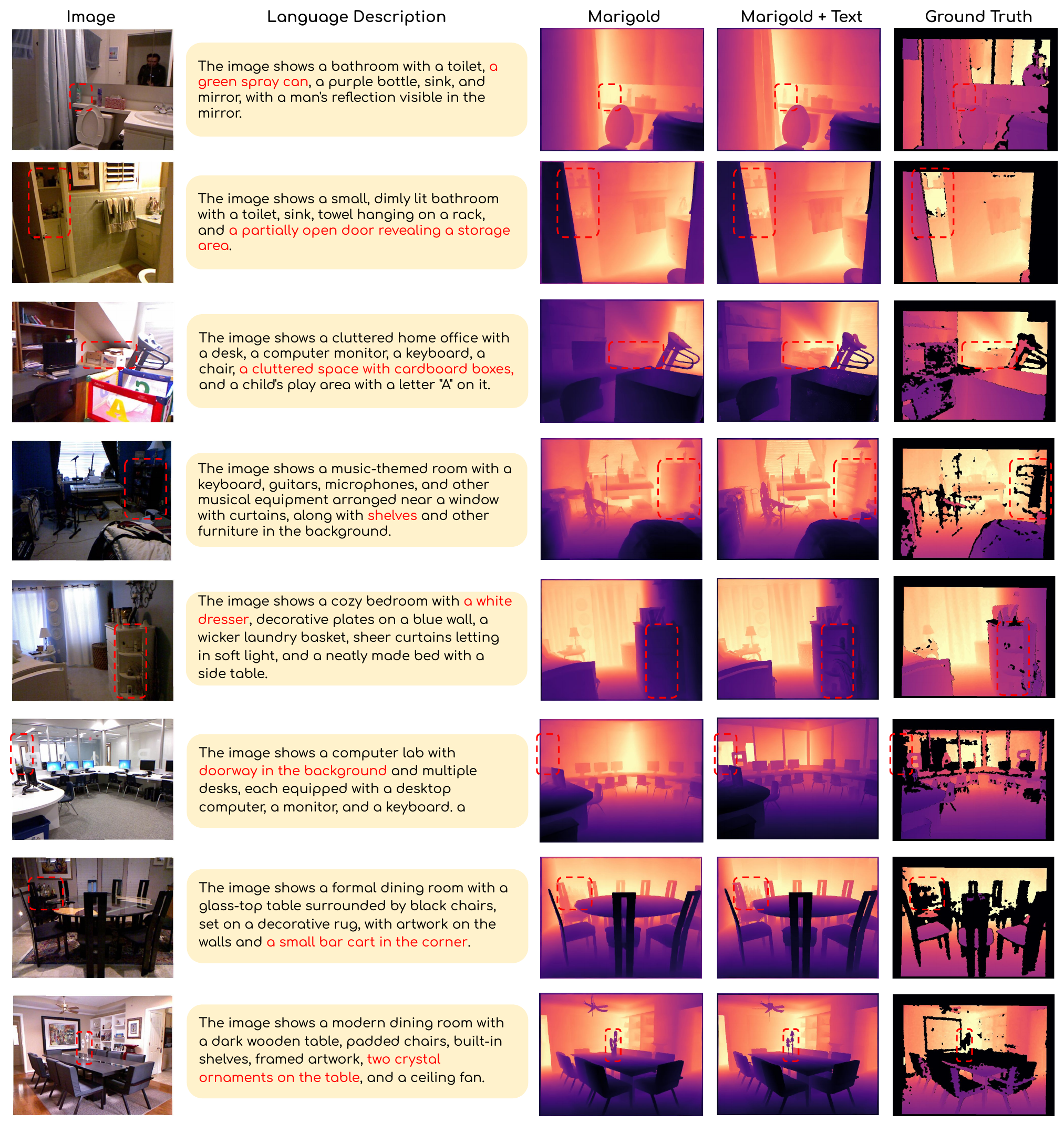}   \caption{\textbf{Additional visualization on NYUv2.} Compared to Marigold, our PriorDiffusion demonstrates better depth prediction, particularly for instances specified in the language description (highlighted in red text and marked with red boxes). The language description effectively guides the model's attention to relevant regions, especially those easily overlooked by visual cues due to a small size or a transparent texture, such as ``a green spray can'' in the 1st row and ``two crystal ornaments'' in the last row. It also improves perception under challenging visual conditions, like ``shelves'' in the 4th row and ``a white dresser'' in the 5th row, both of which are under poor illumination and are difficult to tell from visual solely. Additionally, it supports complex reasoning about scene layouts that might be misinterpreted from visual cues alone, such as ``a doorway in the background'' in the 6th row. Furthermore, it provides critical context for partially observed or occluded objects, such as ``a partially open door revealing a storage area'' in the 2nd row and ``a small bar cart in the corner'' in the 7th row.
  }
    \label{fig:vis_nyu_supp}
    \vspace{-0.3cm}
\end{figure*}

\begin{figure*}[t!]
  \centering
    \includegraphics[width=1.0\textwidth]{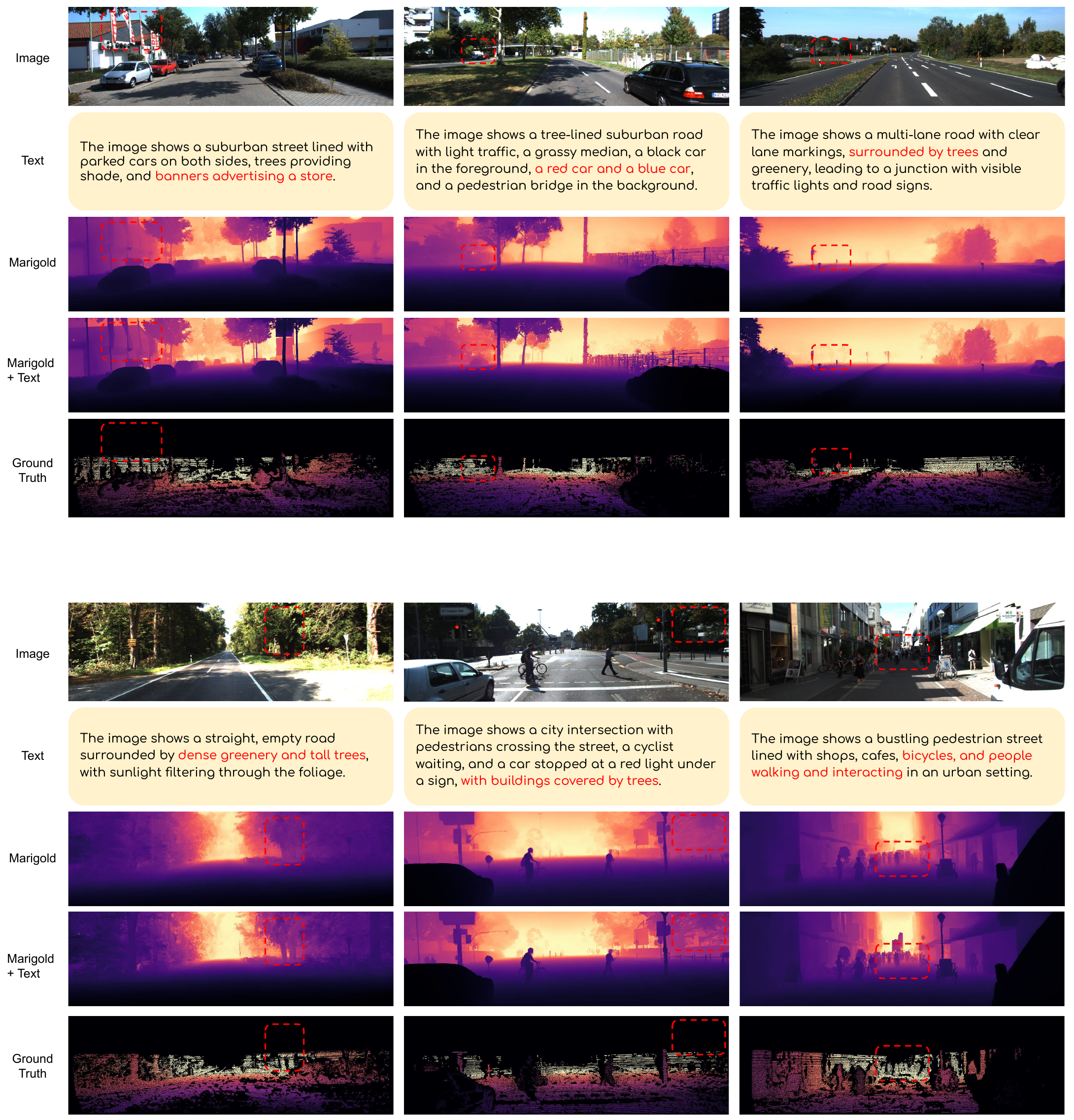}   \caption{\textbf{Additional visualization on KITTI.} Compared to Marigold, our PriorDiffusion demonstrates superior depth prediction, particularly for instances specified in the language description (highlighted in red text and marked with red boxes). The language description effectively guides the model's attention to relevant regions that might otherwise be overlooked due to their small size or subtle visual cues. Examples include ``banners advertising a store'' in the upper 1st column and ``a red car and a blue car'' in the upper 2nd column. Additionally, it enhances perception in complex scenes featuring multiple objects or intricate surfaces. For instance, it accurately captures ``surrounded by trees'' in the lower 1st column, ``dense greenery and tall trees'' in the upper 3rd column, and ``bicycles, and people walking and interacting'' in the lower 3rd column. Furthermore, the language descriptions provide essential context for partially observed or occluded objects, such as ``buildings covered by trees'' in the lower 2nd column.}
    \label{fig:vis_kitti_supp}
    \vspace{-0.3cm}
\end{figure*}

We provide additional visualization and analysis for indoor scenes in Figure~\ref{fig:vis_nyu_supp} and outdoor scenes in Figure~\ref{fig:vis_kitti_supp}. We use several samples in the NYUv2~\cite{NYU-Depth-V2} and KITTI~\cite{KITTI,KITTI_1} dataset across diverse types of scenes. We have provided examples in captions under each figure. These visualizations demonstrate that leveraging the language prior within the text-to-image diffusion model enhances the model's ability to understand the geometric characteristics of the specified regions and objects. It shows that language plays a critical role in guiding the model's attention to relevant regions and providing context for improved depth prediction. It highlights subtle or easily overlooked details, such as small objects or instances, and enhances the perception of complex scenes with multiple objects or intricate surfaces. Additionally, language descriptions offer an essential context for partially observed or occluded objects, enabling the model to infer details that visual cues alone might miss. By integrating language prior, the model achieves a more comprehensive and accurate understanding of scenes, especially in challenging scenarios.

\clearpage
\subsection*{C.2. Ablation for prompts to generate text}
To study the effect of different prompts for text generation, here we use Marigold~\cite{Marigold} and the visual question-answering model LLaVA v1.6 Mistral~\cite{liu2023improved} to generate one text for each training image and testing image. The prompt we use should elicit responses that capture essential details, including the positioning of objects, their interactions, and notable features that influence monocular depth estimation. We generate different prompts using ChatGPT 4o~\cite{openai2024chatgpt4}, with the prompt:

\textit{``Generate prompt for a vision-language model to generate language description for each given image in one sentence. The prompt we use needs to elicit responses that include essential details, such as the positioning of objects, their interactions, and notable features that may impact depth estimation." }

We present the results in Table~\ref{tab:prompts_ablation_supp}, showcasing various language descriptions generated by LLaVA under different prompts. While the performances exhibit some variation, they remain consistently comparable. This demonstrates that diffusion-based depth estimators can be enhanced as long as they are provided with meaningful language descriptions of 3D scenes that resemble natural human descriptions.

\begin{table}[t]
\centering
\begin{adjustbox}{width=0.5\textwidth}
\begin{tabular}{@{}lcccccccc@{}}
\toprule
\multirow{2}{*}{Method} & \multicolumn{2}{c}{NYUv2} & & \multicolumn{2}{c}{KITTI} & & \multicolumn{2}{c}{ETH3D} \\
\cmidrule{2-3} \cmidrule{5-6} \cmidrule{8-9}
& $\delta_1$↑ & AbsRel↓ & & $\delta_1$↑ & AbsRel↓ & & $\delta_1$↑ & AbsRel↓ \\
\midrule
Marigold & 95.7	& 6.1	& &89.7	& 10.4	& & 95.4	& 6.9\\
``An image" &95.7  &6.1     &&89.8  &10.7   &&95.1  &6.8   \\
Template A &95.8  &6.0     &&90.3  &10.6   &&95.5  &6.4   \\
Template B &95.7  &6.1     &&90.5  &10.7   &&95.6  &6.5   \\
Template C &\textbf{95.9}  &6.0     &&90.2  &10.6   &&95.3  &6.6   \\
Template D &95.6  &\textbf{5.8}     &&90.4  &10.5   &&95.4  &6.4   \\
Template E &95.8  &5.9     &&90.3  &10.7   &&95.6  &6.7   \\
Template F &\textbf{95.9}  &6.0     &&90.5  &10.5   &&95.4  &6.6   \\
Template G &95.8  &6.1     &&90.3  &10.6   &&95.5  &\textbf{6.3}   \\
Template H &95.7  &5.9     &&90.2  &10.7   &&\textbf{95.7}  &\textbf{6.3}   \\
Template I &95.8  &6.1    &&\textbf{90.6}  &10.5   &&95.4  &6.6   \\
Template J &95.8  &6.0     &&90.5  &10.6   &&95.6  &6.4   \\
Template K &95.7  &\textbf{5.8}     &&90.4  &10.5   &&95.5  &\textbf{6.3}   \\
Ours  & \textbf{95.9} &5.9 & & \textbf{90.6} & \textbf{10.4} & & \textbf{95.7} &6.5\\
\bottomrule
\end{tabular}
\end{adjustbox}
\begin{tablenotes}
\footnotesize
\item[]Template A: ``Describe the image in one sentence. Explain the image by identifying key objects and their distances from the viewpoint, noting any perspective lines or depth cues that indicate the three-dimensional structure of the scene.''\\

\item[]Template B: ``Describe the image in one sentence. Describe the image by detailing the foreground, midground, and background objects, emphasizing their relative distances and spatial positioning within the scene.'' \\

\item[]Template C: ``Describe the image in one sentence. Describe the image by detailing the foreground, midground, and background objects, emphasizing their relative distances and spatial positioning within the scene.'' \\

\item[]Template D: ``Describe the image in one sentence. Provide an in-depth description of the image, focusing on the scale and depth of each visible object and how they overlap or are spaced from one another.'' \\

\item[]Template E: ``Describe the image in one sentence. Analyze the image by discussing the size and arrangement of objects, their positions relative to one another, and any changes in texture or clarity that indicate varying depths across the scene.'' \\

\item[]Template F: ``Describe the image in one sentence. Describe the scene with attention to depth, specifying which elements appear closer or farther from the viewer and how shadows or lighting contribute to the perception of depth.'' \\

\item[]Template G: ``Describe the image in one sentence. Highlight the depth relationships in the image by describing which objects are in the foreground, which are in the background, and how their relative sizes help convey distance.'' \\

\item[]Template H: ``Describe the image in one sentence. Focus on any natural or man-made structures in the image and describe how their orientation and placement give a sense of depth or perspective.'' \\

\item[]Template I: ``Describe the image in one sentence. Describe how elements like roads, pathways, or fences create leading lines that guide the viewer's eye into the depth of the scene.'' \\

\item[]Template J: ``Describe the image in one sentence. Explain how differences in lighting or shadowing in the image indicate which parts are nearer or further away from the observer.'' \\

\item[]Template K: ``Describe the image in one sentence. Analyze the spatial arrangement of the main objects and describe any overlapping or occlusion that suggests depth relationships between them. ''

\item[]Ours: ``Describe the image in one sentence, assuming it's a real-world image, pay more attention to objects, their spatial relationships, and the overall layout.'' \\

\end{tablenotes}
\caption{\textbf{Ablation for prompts to generate language description.} Prompts are used to prompt LLaVA to generate language descriptions for each image. While the performances among different prompts may vary, they remain consistently comparable, as long as they are meaningful and mimic human descriptions. Marigold in the first row is trained without text.}
\label{tab:prompts_ablation_supp}
\vspace{-0.5cm}
\end{table}

\subsection*{C.3. Different denoising steps}
\label{sec:denoising_steps}
As demonstrated in Figure~\ref{fig:denoising_steps}, we Marigold baseline and Marigold with both training text and inference text, with different denoising steps during inference. The performances are shown in Table~\ref{tab:denoising_steps}. With more denoising steps, performance gradually improves. Integrating text consistently outperforms the baseline across various denoising steps, converging in just 10 steps, while the baseline requires 25 steps. This suggests that the language can speed up the denoising process and accelerating convergence.

\begin{table}[t!]
\centering
\begin{adjustbox}{width=0.5\textwidth}
\begin{tabular}{@{}lcccccccccccc@{}}
\toprule
\multirow{2}{*}{Steps} & \multicolumn{2}{c}{NYUv2} & & \multicolumn{2}{c}{KITTI} & & \multicolumn{2}{c}{ETH3D} & & \multicolumn{2}{c}{ScanNet} \\
\cmidrule{2-3} \cmidrule{5-6} \cmidrule{8-9} \cmidrule{11-12}
& $\delta_1$↑ & AbsRel↓ & & $\delta_1$↑ & AbsRel↓ & & $\delta_1$↑ & AbsRel↓ & & $\delta_1$↑ & AbsRel↓ \\
\midrule
\multicolumn{12}{c}{\textbf{Marigold}} \\
1 & 48.8 & 33.8 & & 25.2 & 59.1 & & 50.1 & 37.5 & & 60.3 & 25.7 \\
2 & 78.7 & 16.9 & & 72.2 & 18.0 & & 86.0 & 12.7 & & 72.1 & 19.2 \\
3 & 92.6 & 8.2 & & 77.7 & 15.3 & & 92.0 & 8.6 & & 87.1 & 11.4 \\
4 & 94.2 & 7.0 & & 80.4 & 14.1 & & 93.6 & 7.6 & & 90.7 & 9.3 \\
5 & 94.7 & 6.7 & & 82.4 & 13.4 & & 94.3 & 7.3 & & 91.9 & 8.6 \\
10 & 95.1 & 6.4 & & 86.5 & 11.9 & & 94.9 & 6.9 & & 93.2 & 7.8 \\
15 & 95.1 & 6.4 & & 87.4 & 11.6 & & 95.0 & 6.9 & & 93.4 & 7.7 \\
20 & 95.2 & 6.3 & & 88.0 & 11.4 & & 95.0 & 6.9 & & 93.7 & 7.5 \\
25 & 95.3 & 6.3 & & 88.2 & 11.3 & & 95.0 & 6.9 & & 93.7 & 7.5 \\
50 & 95.3 & 6.3 & & 88.5 & 11.2 & & 95.0 & 6.9 & & 93.8 & 7.5 \\
\midrule
\multicolumn{12}{c}{\textbf{Marigold + Text (Training \& Inference)}} \\
1 & 48.8 & 33.8 & & 25.2 & 59.1 & & 50.1 & 37.5 & & 60.3 & 25.7 \\
2 & 83.2 & 14.1 & & 74.7 & 16.9 & & 86.5 & 12.0 & & 75.0 & 17.8 \\
3 & 94.3 & 7.1 & & 81.2 & 13.9 & & 93.0 & 7.7 & & 90.6 & 9.5 \\
4 & 95.5 & 6.3 & & 84.7 & 12.4 & & 94.7 & 6.9 & & 93.5 & 7.8 \\
5 & 95.7 & 6.0 & & 87.0 & 11.6 & & 95.3 & 6.6 & & 91.9 & 8.6 \\
10 & 96.0 & 5.9 & & 89.9 & 10.5 & & 95.7 & 6.5 & & 94.9 & 6.8 \\
15 & 96.0 & 5.9 & & 90.1 & 10.4 & & 95.8 & 6.5 & & 94.9 & 6.8 \\
20 & 96.0 & 5.9 & & 90.3 & 10.4 & & 95.8 & 6.5 & & 94.9 & 6.7 \\
25 & 96.0 & 5.9 & & 90.3 & 10.4 & & 95.8 & 6.5 & & 94.9 & 6.7 \\
50 & 96.0 & 5.9 & & 90.3 & 10.4 & & 95.8 & 6.5 & & 94.9 & 6.7 \\
\bottomrule
\end{tabular}
\end{adjustbox}
\caption{\textbf{Performance for different denoising steps.}  Integrating text consistently outperforms the baseline across various denoising steps, with a significantly faster convergence speed for the diffusion process.}
\label{tab:denoising_steps}
\vspace{-0.3cm}
\end{table}

\noindent
\subsection*{C.4. Template Prompt Comparison}
\label{sec:Template_prompt_comparison.}
When training with text, we also consider inference scenarios where user-provided descriptions are unavailable. We therefore explore whether the model can still perform comparably by using either a blank input or standardized template prompts. Specifically, we use the Marigold model trained with text and evaluate the effect of several predefined prompts—ranging from simple ones such as blank string ``'', simple words like ``An image'', to more descriptive ones like ``A complex 3D scene with varying objects at different distances.'', as inputs to the diffusion-based depth estimator to maintain its performance.

As presented in Table~\ref{tab:template_prompt}, these template prompts help preserve the model’s performance when explicit language input is not feasible. The results show that the model achieves comparable, or even better, performance than the Marigold baseline when using fixed prompts instead of user-provided text. This finding suggests that, even when user-provided descriptions are unavailable, incorporating language during training itself might enhance the depth estimator’s generalization and overall performance.

\begin{table}[t]
\centering
\begin{adjustbox}{width=0.5\textwidth}
\begin{tabular}{@{}lcccccccc@{}}
\toprule
\multirow{2}{*}{Method} & \multicolumn{2}{c}{NYUv2} & & \multicolumn{2}{c}{KITTI} & & \multicolumn{2}{c}{ETH3D} \\
\cmidrule{2-3} \cmidrule{5-6} \cmidrule{8-9}
& $\delta_1$↑ & AbsRel↓ & & $\delta_1$↑ & AbsRel↓ & & $\delta_1$↑ & AbsRel↓ \\
\midrule
Blank text input & 95.5 & 6.2 && 89.3 & 10.9 && 95.0 & 6.9 \\
``An image'' & 95.7 & 6.1 && 89.8 & 10.7 && 95.1 & 6.8 \\
Template A & \textbf{95.8} & \textbf{6.0} && \textbf{89.9} & 10.7 && 95.3 & 6.8 \\
Template B & 95.7 & 6.1 && 89.8 & 10.7 && 95.2 & 6.8 \\
Template C & \textbf{95.8} & 6.1 && 89.8 & 10.7 && \textbf{95.3} & \textbf{6.8} \\
Marigold$^*$ & 95.7 & 6.1 && 89.7 & 10.7 && \textbf{95.4} & 6.9 \\
\hline
With text input & 95.9 & 5.9 & & 90.6 & 10.4 & & 95.7 & 6.5 \\
\bottomrule
\end{tabular}
\end{adjustbox}
\begin{tablenotes}
\scriptsize
\item[a]Template A: `A complex 3D scene with 
varing objects at different distances." \\\item[b]Template B: `A structured environment with intricate patterns and designs that create depth and guide the eye through various focal points."\\
\item[c]Template C: `An elaborate scene with overlapping objects that create a sense of distance and spatial hierarchy within the environment." \\
\end{tablenotes}
\caption{\textbf{Inference using fixed template text input.} The results show that the model achieves comparable, or even better, performance than the Marigold baseline when using fixed prompts instead of user-provided text. This finding suggests that, even when user-provided descriptions are unavailable, incorporating language during training itself might enhance the depth estimator’s generalization and overall performance.}
\label{tab:template_prompt}
\vspace{-0.5cm}
\end{table}

\subsection*{C.5. Ablation on the Number of Text Captions per Image} 

We test Marigold's performance with different numbers of captions provided for each image, to test whether increasing the number of text provided during training and improve the performance. To generate text descriptions for images, for training images, we use two different version of LLaVA v1.6~\cite{liu2023improved}, Mistral and Vicuna, each with 5 different prompts, to generate 10 text descriptions for each training image: 

\begin{itemize}
    \item \textit{``Describe the image in one sentence, assuming it's a real-world image.''}
    \item \textit{``Provide a one-sentence description of the image, pay attention to object type, assuming it's a real-world image.''}
    \item \textit{``Capture the essence of the image in a single sentence, pay attention to object relationship, assuming it's a real-world image.''}
    \item \textit{``Condense the image description into one sentence, pay attention to object size, assuming it's a real-world image.''}
    \item \textit{``Express the image in just one sentence, pay attention to the overall layout, assuming it's a real-world image.''}
\end{itemize}

For testing images, we use LLaVA v1.6 Mistral, then prompt this model with: 

\begin{itemize}
\item \textit{``Describe the image in one sentence, assuming it's a real-world image. Pay close attention to objects, their spatial relationships, and the overall layout." }
\end{itemize}

This prompt encourages generated responses to include essential details for depth estimation, such as the positioning of objects, their interactions, and notable features that may impact depth estimation. Note that all training data we use are synthetic data, and by emphasizing “assuming it's a real-world image,” we ensure that the descriptions align with the types of inputs and scenarios the model will encounter.

When multiple captions are available for each image, one is randomly selected during training. Shown in Table~\ref{tab:number_caption}, as the number of captions increases, performance saturates and the improvement is marginal. One possible explanation is that different captions provides similar descriptions of the scene in terms of attributes essential for depth estimation, such as object types, sizes, spatial relationships, and scene structure, since those captions are prompted to cover all those essential details. Generating additional captions does not introduce additional information, thus leading to only marginal improvements in monocular depth estimation accuracy.

\begin{table}[t]
\centering
\begin{adjustbox}{width=0.5\textwidth}
\begin{tabular}{@{}lcccccccc@{}}
\toprule
\multirow{2}{*}{Method} & \multicolumn{2}{c}{NYUv2} & & \multicolumn{2}{c}{KITTI} & & \multicolumn{2}{c}{ETH3D} \\
\cmidrule{2-3} \cmidrule{5-6} \cmidrule{8-9}
& $\delta_1$↑ & AbsRel↓ & & $\delta_1$↑ & AbsRel↓ & & $\delta_1$↑ & AbsRel↓ \\
\midrule
1 Caption Per Image & 95.9 & 5.9 & & 90.6 & 10.4 & & 95.7 & 6.5 \\
2 Captions Per Image & 95.9 & 5.9 & & 90.5 & 10.4 & & 95.7 & 6.5 \\
5 Captions Per Image & 96.0 & 5.9 & & 90.4 & 10.4 & & 95.8 & 6.5 \\
10 Captions Per Image & 96.0 & 5.9 & & 90.3 & 10.4 & & 95.8 & 6.5 \\
\end{tabular}
\end{adjustbox}

\caption{\textbf{Training with different numbers of text captions per image.} For images annotated with multiple captions, a caption is randomly sampled for each training iteration. While adding more captions initially slightly improves performance, the benefit quickly saturates, yielding only minor gains beyond a certain point.}
\label{tab:number_caption}
\vspace{-0.5cm}
\end{table}